\documentclass[12pt]{article}

\usepackage{graphicx}
\usepackage{xcolor}
\usepackage{scicite}
\usepackage{soul}
\usepackage{array}
\usepackage{lineno}
\usepackage{times}
\usepackage[font={sf,footnotesize}]{caption}

\newenvironment{sciabstract}{%
\begin{quote} \bf}
{\end{quote}}

\title{Embodying Physical Computing into Soft Robots}

\author
{Jun Wang,$^{+}$$^{\ast}$ Ziyang Zhou,$^{+}$ Ardalan Kahak,$^{+}$ and Suyi Li$^{\ast}$\\
\\
\normalsize{Department of Mechanical Engineering, Virginia Tech}\\
\normalsize{635 Prices Fork Road, Blacksburg, VA 24061, USA}\\
\\
\normalsize{$^{+}$These authors contributed equally}\\
\normalsize{$^\ast$To whom correspondence should be addressed: \texttt{junw@vt.edu}}
}

\date{}

\begin{document} 

\baselineskip 24pt


\maketitle 


\begin{sciabstract}
\textit{Summary:} Embodying physical computing, analog or algorithmic, presents an avenue for the next generation of entirely soft and intelligent robots.

\medskip 

\textit{Abstract:} Softening and onboarding computers and controllers is one of the final frontiers in soft robotics towards their robustness and intelligence for everyday use.  In this regard, embodying soft and physical computing presents exciting potential.  Physical computing seeks to encode inputs into a mechanical computing kernel and leverage the internal interactions among this kernel's constituent elements to compute the output. Moreover, such input-to-output evolution can be re-programmable. This perspective paper proposes a framework for embodying physical computing into soft robots and discusses three unique strategies in the literature: analog oscillators, physical reservoir computing, and physical algorithmic computing. These embodied computers enable the soft robot to perform complex behaviors that would otherwise require CMOS-based electronics --- including coordinated locomotion with obstacle avoidance, payload weight and orientation classification, and programmable operation based on logical rules. This paper will detail the working principles of these embodied physical computing methods, survey the current state-of-the-art, and present a perspective for future development. 
\end{sciabstract}

\section*{Introduction}
The dream of creating entirely soft, versatile, and capable robots --- akin to the octopus --- has long inspired scientists and engineers. We have witnessed significant progress in soft actuation \cite{el2020soft, rus2015design}, sensing \cite{wang2018toward}, and power \cite{aubin2022towards}, enabling these robots to operate in a wide range of challenging environments, from deep within our own bodies \cite{cianchetti2018biomedical} to the far bottom of the Mariana trench \cite{li2021self}.
Yet, softening and onboarding computers and controllers remain a major challenge and present one of the final frontiers towards robust and intelligent soft robots suitable for everyday use. 
In this regard, roboticists have long recognized that the inherent material softness can facilitate and simplify control, and many innovative strategies have been explored. 
For example, soft and rotating legs can naturally accommodate uneven surfaces and large obstacles, allowing the robot to traverse challenging terrains without complex controls like in the quadrupeds~\cite{saranli2001rhex}. Soft curling tentacles can wrap and entangle themselves around objects with widely different shapes, thus manipulating them with a simple global pressure input~\cite{becker2022active}. 
Such softness-facilitated control is sometimes referred to as \textcolor{black}{``intelligence by mechanics''} \cite{blickhan2007intelligence} or ``morphological computation'' \cite{hauser2011towards, muller2017morphological}.
They offer exciting potential, but frequently lack the sophistication and (re-)~programmability available from the more universal controllers based on digital computation.

In parallel with the advancements in soft robotics \textcolor{black}{(and partly inspired by the need for soft robotic computing and control)}, there is also growing interest in CMOS-free physical computers~\cite {chen2025advances, alu2025roadmap, yasuda2021mechanical, zangeneh2021analogue, qian2025guidance}. This emerging paradigm seeks to \textcolor{black}{\bf encode physical inputs} into a mechanical construct (or \textcolor{black}{\bf kernel}) --- for example, \textcolor{black}{in the form of deformation} ~\cite{mei2023memory, song2019additively, zhang2023meta, liu2023cellular, ducarme2025exotic}, fluid flow \cite{rajappan2022logic, el2020pressure}, thermal heat flux \cite{chen2024thermal}, or waves~\cite{mousa2024parallel, dorin2024embodiment, bilal2017bistable, silva2014performing, tzarouchis2025programmable} --- and leverage the internal interactions among the kernel's constituent elements to process these inputs according to a \textcolor{black}{\bf programmed evolution}. The resulting output typically remains in the same physical domain as the input so that it can be easily \textcolor{black}{\bf decoded} and interpreted. 
\textcolor{black}{The paths to physical computing are diverse: One can use acoustic waves to solve differential equations \cite{silva2014performing, zangeneh2018performing}, re-purpose mechanical vibrations like neuron signals to \hl{perform machine learning tasks} \cite{hauser2011towards, hauser2012role, louvet2025reprogrammable}, or construct mechanical logic gates and physical circuitry to perform algorithmic operations \cite{el2022mechanical}.} 
Overall, the idea of performing computation without CMOS electronics could benefit us with higher energy efficiency \textcolor{black}{\cite{zolfagharinejad2025analogue}}, parallelization \textcolor{black}{\cite{mousa2024parallel}}, and resiliency against adversarial working conditions.

Therefore, there is a tremendous opportunity to introduce physical computing into the field of soft robotics. That is, one could construct a physical computer out of soft materials and integrate it with soft sensors and actuators. Such integration can lead to a new class of entirely soft computation and control methods with flexibility, robustness, and programmability for more sophisticated tasks. As a result, we have witnessed a rapid emergence of soft robots with integrated and embodied physical computers over the past several years. 
\textcolor{black}{And these physically computing robots have become an important part of recent reviews that offer a birds-eye overview on embodied intelligence (or mechanical intelligence, physical control) in robotics \cite{chen2025physical, chen2025advancing, mengaldo2022concise, milana2025physical}.}

\textcolor{black}{On the other hand, we believe a separate, deeper dive into physical computing in soft robots can benefit the research community. Specifically, we aim to define physical computing using a rigorous framework, including encoding, decoding, and a (re-)programmable computing kernel, and build upon this definition to categorize physical computing into two distinct types: analog and algorithmic (more in the following Section 1). In this way, we can dissect and re-examine soft robotics through a different lens. We will also have a more systematic framework to introduce new physical computing concepts from other disciplines to robotics.}

\textcolor{black}{Therefore, this perspective paper will first establish a more formal framework for physical computing, then survey the different analog and algorithmic computers implemented in the soft robots, and discuss challenges and future directions in the end.}

\section{What is physical computing (and what is not)?}
Before surveying soft robots with embodied computing, we should first clarify the definition of a physical computer. In the robotics literature, the scope of ``computing'' and ``intelligence'' has been quite broad and occasionally conflicting. We certainly do not intend to propose a new definition that everyone would agree upon. Instead, we would like to highlight a few key ingredients of the physical computer to anchor the scope of this particular paper.  

\begin{figure}[t!]
    \centering
    \includegraphics[scale=0.85]{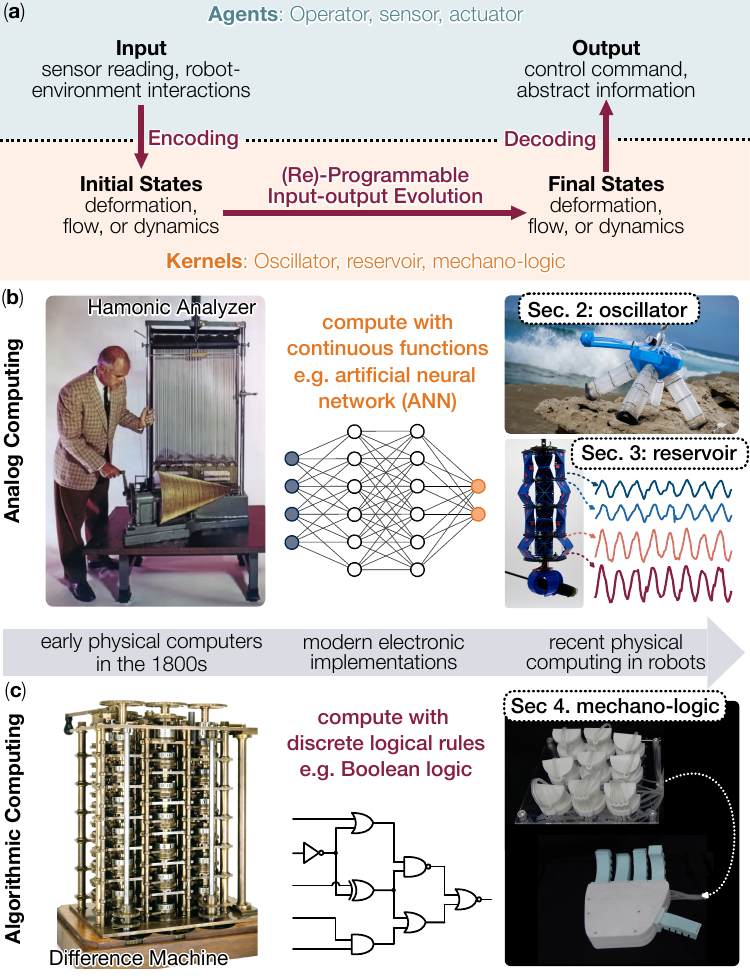}
    \caption{\textbf{The architecture and categorization of physical computing}. (a) \hl{The computing architecture adapted in this perspective includes input encoding, output decoding, and programmable input-output evolution.} 
    \textcolor{black}{
(b) \textbf{Analog computing}: The Harmonic Analyzer is an elegant example of analog computing from the 1800s (photo credit to the Nimitz Library, United States Naval Academy). In the modern electronic computing paradigm, Artificial Neural Networks also operate on analog (continuous) data. We will discuss two types of physical analog computing in soft robots: oscillator and reservoir. For example, an electronics-free legged robot uses an analog oscillator to walk \cite{drotman2021electronics} (photo credit to David Baillot, Jacobs School of Engineering, UC San Diego), and a modular manipulator uses embodied reservoir computing to classify payloads (image adapted from \cite{wang2025re} CC BY 4.0).
(c) \textbf{Algorithmic computing}: The Difference Machine is one of the first algorithmic computers (photo credits to Science Museum London, CC Attribution-SA 2.0). Modern CMOS-based computers are built exclusively on algorithmic Boolean logic.  Here, we survey how mechanical logic is implemented in soft robots. For example, a robotic hand operates with fluidic logic control (image adapted from \cite{jiao2024reprogrammable} CC BY 4.0).
}}
    \label{fig:bigpic}
\end{figure}

To this end, we propose that physical computing should involve two domains: one is the \emph{agents} using the computer. They could be a human operator, but in this study, they typically refer to the non-computing parts of a soft robot, including sensors, actuators, and power supply. The agents will have an ``input'' that they wish to use, and expect an ``output'' from the computing. The second domain is the \emph{kernel}, where the physical interaction between its constituent components embodies the computing program. Under this formalism, a complete physical computer should 1) have a mechanism to \emph{encode} inputs from the agents into the computing kernel and \emph{decode} the outputs correspondingly, and 2) have a mechanism to \emph{program} (i.e., design and configure) the evolution from the input to the output in the computing kernel (Figure~\ref{fig:bigpic}a).

Therefore, \textit{in this paper, computing does not exist without encoding, decoding, and programming} \cite{horsman2014does}. Under this formalism, some nonconventional and innovative computing paradigms in the robotics field, such as the aforementioned ``morphological computation,'' are beyond our scope. 
\textcolor{black}{Morphological computation generally refers to the idea that a robot body's shape, deformation, and dynamics can perform part of the ``computation'' needed for control. Under this paradigm, ``computation'' can be quite diverse --- it can be storing and releasing energy periodically to stabilize locomotion (e.g., passive walker \cite{mcgeer1990passive}), or conforming to complex objects to assist manipulation (e.g., vacuum jamming gripper \cite{brown2010universal}), or pre-processing sensory data to assist perception (e.g., bat ear that mechanically process the incoming sound waves to assist object localization \cite{reijniers2010morphology}). 
Therefore, the physical computing defined in this paper can be an example of morphological computation, but it has a more structured definition. That is, many morphological computation examples will not be considered as physically computing in this paper because they do not have the ``encoding-kernel evolution-decoding'' architecture, and they are not reprogrammable.}

On the other hand, a mechanical construct --- e.g., architected materials or soft robotic body --- that can incorporate encoding, decoding, and programming would meet the necessary condition to function as a physical computer.
Moreover, based on these definitions, we will adopt the theory from Jaeger et al., and categorize physical computers into two sets (Figure \ref{fig:bigpic}b) \cite{jaeger2021towards, jaeger2023toward}.

One is \textit{analog}, where input and output signals are continuous, and the evolution from input to output is governed by smooth (and frequently physics-based) functions. Albert Michelson’s harmonic analyzer \cite{hammack2014albert} and our human neural system are classical examples of analog computers. In soft robots, this can be accomplished by exploiting their bodies' nonlinear dynamic responses for physical reservoir computing. 
The other type of physical computer is the \textit{algorithmic}, where the input and output take a discrete format, and the evolution from input to output is programmed via a set of abstract logical rules. Charles Babbage’s difference machine \cite{swade2001difference} and our omnipresent CMOS-based computing chips are classical examples of an algorithmic computer. In soft robots, this can be accomplished by, for example, an assembly of mechanical Boolean gates featuring elastic bistability (i.e., mechanical logic gates).  \textcolor{black}{Table \ref{tab:summary} summaries and compares the different computing approaches from this perspective.}

\begin{table}[t]
\centering
\small
\textcolor{black}{
\textsf{
\begin{tabular}{ || m{1.25cm} | m{1.75cm}| m{6cm} | m{4.5cm} ||} 
  \hline
 section & computing approach & working principle  & suitable tasks \\ 
  \hline
  \hline
  2 & \underline{analog} oscillator & use nonlinear mechanics to generate continuous and rhythmic signals (and motions)  & locomotion gait sequencing \\ 
  \hline
  3 & \underline{analog} reservoir & extract nonlinear input-output mapping or complex information from high-dimensional body dynamics & classification, multi-modal sensing, and locomotion control  \\
  \hline
  4 & \underline{algorithmic} computer & use mechanical logic gates to conduct discrete operations according to abstract logical rules  & user interfacing and control, behavior switching \\
  \hline
\end{tabular}
}
}
\caption{Summary of physical computing approaches discussed in this perspective.} \label{tab:summary}
\end{table}

\subsection*{Responsiveness and adaptation do not necessarily mean computation}

It is worth highlighting that many soft robots use responsive materials to interact with their surrounding environment and achieve adaptive behaviors.  However, they do not necessarily compute according to the above-mentioned definition. 

Materials are considered ``responsive'' or ``active'' if they can change their shape or constitutive properties in response to external stimuli, such as temperature \cite{zhang2021wirelessly}, heat flux \cite{he2023modular}, electric field \cite{wang2023versatile}, magnetic field \cite{zhang2021wirelessly, fan2020reconfigurable}, light \cite{he2023modular, wang20233d}, and humidity \cite{luo2023autonomous}. 
They were initially introduced to soft robotics as artificial muscles. Shape memory alloys (SMAs) have been widely utilized in soft robotics since their inception \cite{rodrigue2017overview}. Dielectric elastomer is another example \cite{guo2021review}, and some liquid format dielectric materials can generate very high output forces to create jellyfish-like soft robots \cite{wang2023versatile}. A programmable electrothermal actuator using silver nanowires (AgNW) can enable a robot to crawl \cite{wu2023caterpillar}. One can also harvest responsive materials from nature, such as the self-drilling seed carrier made from white oak tissue, which can autonomously burrow by exploiting ambient humidity cycles \cite{luo2023autonomous}. (Interested readers can refer to the excellent reviews in \cite{zhao2022stimuli, shen2020stimuli} for a comprehensive survey of responsive materials used for robotic actuation.)

As responsive materials continue to evolve, researchers are beginning to explore how they can be strategically embedded in soft robotic bodies to facilitate and simplify control. For example, untethered robots with responsive materials can achieve simple and remote operation \cite{boyvat2021remote}, thereby reducing the associated control and computational complexity. Examples like this include miniature magnetic shape-programmable robots \cite{zhang2021wirelessly, fan2020reconfigurable}, m-PDMS (magnetic particle-doped polydimethylsiloxane) robots \cite{ke2024synergistical, soon2023pangolin, wang2022adaptive}, photoresponsive LCE robots \cite{wang20233d}, and piezoelectric polyvinylidene fluoride (PVDF) robots \cite{mu2023spiral}. By integrating different types of responsive materials in one body, simple computational capabilities \cite{he2023modular, he2024modular} can be achieved (e.g., a soft robot that turns toward light only if heat is also present.)

However, although these responsive materials can enable complex tasks without sophisticated controllers (which suggests some intelligence in the mechanical domain, or mechano-intelligence~\cite{sitti2021physical, li2023physical}), they are not considered computing in this study due to a lack of clear mechanisms for input encoding, output decoding, and programmable input-output evolution. Instead, responsive materials can serve as the building blocks of physical computing, and we hope this will become clear as we survey soft robots with physical computing in the following sections.

\section{Analog Oscillators and Rhythmic Motion}


Rhythmic motions here refer to periodic changes in the shape or configuration of a soft structure over time. They are omnipresent in the animal kingdom, such as breathing, heart beating, and in particular, locomotions like walking, swimming, and wing flapping~\cite{dickinson2000animals}. The underpinning mechanisms to generate rhythmic motions are diverse and still active topics of research. Among them, the central pattern generator (CPG) is a unique mechanism that can be considered as a physical computer and thus directly relates to this study. CPG is a self-organized neural circuit that produces rhythmic output from a simple, non-rhythmic input, and the input-output evolution is programmed by the neural circuit's architecture. The central pattern generator makes it possible to achieve and reconfigure complex locomotion gaits with minimal involvement from the brain or local sensory feedback~\cite{mackay2002central, ijspeert2008central}. 

The striking simplicity and capability of the central pattern generator have inspired similar implementations in soft and continuous robots, where an \textit{analog oscillator} --- either electric or mechanical --- is integrated to generate rhythmic deformation from a simple (and typically constant) input to drive locomotion \cite{zhou2023cpg}. Although many of these oscillators are not as complex as CPG's neural circuit, their underlying computing principle are similar. 

\begin{figure}[t!]
    \centering
    \includegraphics[scale=0.85]
    {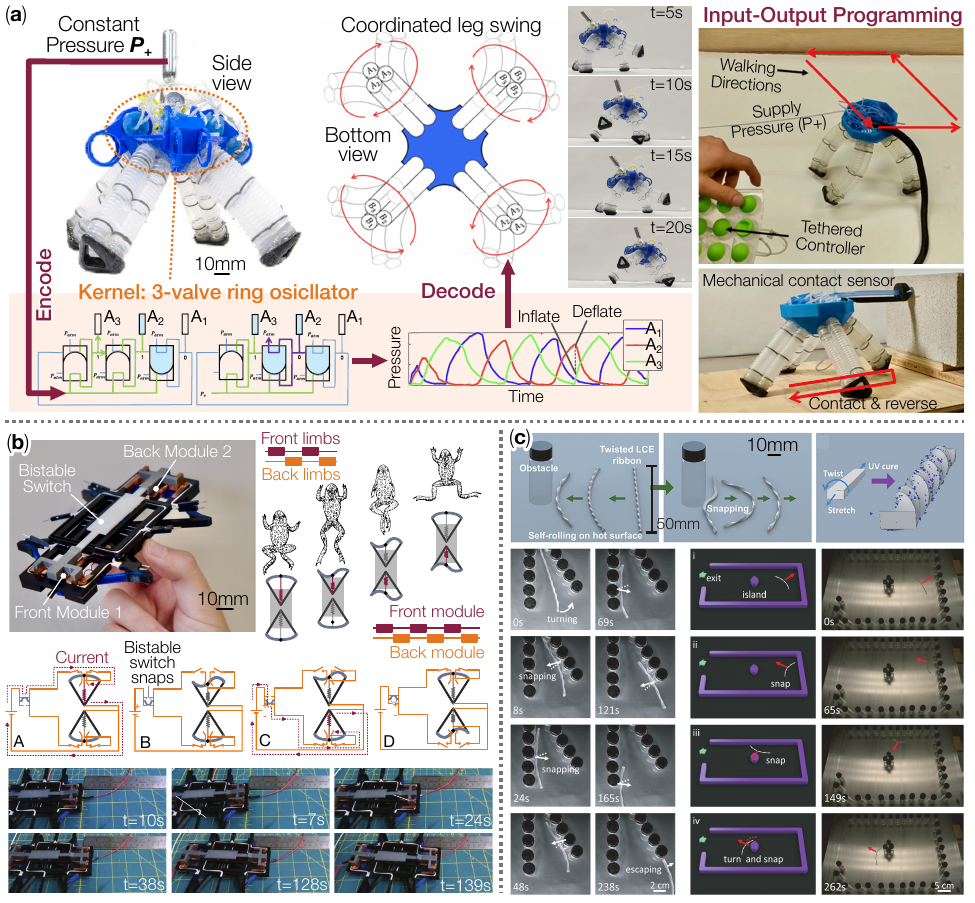}
    \caption{\textbf{Analog oscillators and rhythmic deformation}. (a) Electronics-free pneumatic control: A soft ring-oscillator circuit generates rhythmic leg actuation from a constant pressure input, enabling a quadruped to produce diagonal-couplet walking gaits. A bistable 4/2 switch selects gait direction, and dual oscillators set the phase between leg pairs (image adapted from~\cite{drotman2021electronics} with permission). (b) Controller-free SMA modular robot: A curved mono stable beam and a mechanical slider can convert a single DC power supply into sustained self-oscillation. And a bistable switch can alternate power supply between the front and back modules for out-of-phase deformation and crawling (\textcolor{black}{adapted from \cite{zhou2024self}, CC BY 4.0}). (c) Twisted LCE ribbon robot: The ambient heating drives continuous self-rolling of this robotic structure for locomotion.  When the robot contacts an obstacle, it will store elastic energy and then snap to reverse its direction, enabling autonomous avoidance and maze escape (\textcolor{black}{adapted from \cite{zhao2022twisting}, CC BY-NC-ND 4.0}). Note that all scale bars are approximate.}
    \label{fig:RD}
\end{figure}

\textcolor{black}{\underline{An example of analog oscillator applied to soft robots}}:
In the quadruped robot shown in Figure~\ref{fig:RD}(a), rhythmic and coordinated leg swing motions are generated by an entirely pneumatic ring oscillator~{\cite{drotman2021electronics}}. More specifically: 

\begin{itemize}
    \item \textcolor{black}{\textit{Input encoding}: In this robot, a small pressure tank supplies a constant pressure ($P_+$) to drive the analog oscillator.}

    \item \textcolor{black}{\textit{Kernel}: The oscillator circuit is the kernel. The three soft valves inside this circuit serve as inverters with a built-in delay, and a snap-through membrane alternates between closed and open flow paths, allowing the high-pressure flow to advance around the ring. This mechanism essentially transforms the steady input into a phase-shifted sequence of pressure pulses at the three nodes.}

    \item \textcolor{black}{\textit{Output decoding}: The pulsed output pneumatic pressure from the oscillator flows to the corresponding soft legs, which convert the pressure inflation into mechanical swing motions for walking.}

    \item \textcolor{black}{\textit{Re-programming}: In addition, a soft bistable valve and tethered mechanical controller are added to swap two connections like a latching switch, so triggering the valve can reverse the output pressure pulse sequence, thus reversing the locomotion direction.}
\end{itemize}

Besides the fluidic circuitry, analog oscillation can also be achieved using other physical principles and material selections (Fig. \ref{fig:RD}b-c). For example, one can exploit mechanical instabilities and clever geometric design to generate motion with a constant power input, as seen in twisted LCE ribbons and architected structures that exploit snapping or buckling for autonomous rolling or twisting {\cite{goswami20193d, zhao2022twisting}}. Similarly, beetle-like robots use spiral-shaped PVDF materials to generate mechanical resonance and rhythmic motion for insect-scale and high-speed crawling \cite{yang202088, mu2023spiral,chen2025non}. One can also use thermal or mechanical loops — such as SMA-actuated systems with built-in mechanical switching or microfluidic logic circuits — to generate self-sustained rhythmic actuation without digital controllers {\cite{zhou2024self, wehner2016integrated, kotikian2019untethered}}. 

\textcolor{black}{\underline{Challenges and opportunities of analog oscillators}: Analog oscillators are simple yet robust. They can tightly integrate with the soft robot's body to generate locomotion without additional electronics. However, analog oscillators can suffer from programmability and scalability constraints: Their dynamics are hard-wired into physical design. That is, the oscillator geometry, mechanical architecture, and constitutive material properties fully determine the output frequency and phase pattern. As a result, “programming'' the kernel’s input-output evolution might require re-design rather than a straightforward parameter tuning. In addition, as the number of oscillators increases, fabrication tolerances and material variability can introduce mismatches that degrade synchronization. One can address these limitations by integrating the oscillators with other, more easily programmable components (e.g., combining the oscillator with fluidic logic gates as we show later in Section 4), and using high-precision manufacturing techniques at the smaller physical scale (as we discuss later in Section 5). }

\section{Analog Physical Reservoir Computing}
 
While an analog oscillator provides a promising alternative to micro-controllers for computing and generating rhythmic motions, its information processing capability is largely embedded in its physical architecture. \textcolor{black}{Recent work has shown that architected mechanical and metamaterial-based systems can support multiple motion sequences through controlled switching of actuation frequency~\cite{mousa2024frequency,van2024bio,comoretto2025physical}, rather than through real-time algorithmic control. Nevertheless, the space of attainable behaviors in these systems remains discretely prescribed by design and reconfiguration pathways.} In contrast, there is an emerging notion of informational embodiment, where the combinatorial richness of a soft robot’s body deformation encodes spatiotemporal patterns without relying on symbolic or centralized representations~\cite{pitti2025informational}. This concept bridges soft-body dynamics and computation, offering another pathway toward decentralized, analog computing.

Aligned with this, \textbf{P}hysical \textbf{R}eservoir \textbf{C}omputing (\textbf{PRC}) offers a rigorous framework to formalize the soft robotic body as a computing kernel. In PRC, the body functions as a nonlinear, high-dimensional, and transiently stable dynamical system that maps input streams into distinguishable physical states. In other words, the soft body serves as a ``physical recurrent neural network,'' and its rich dynamics can substitute the digital recurrent neural network for temporal information processing. As we discussed in Section 2, a physical system truly computes only when it is intentionally used to compute abstract functions through defined input encoding and output decoding~\cite{muller2017morphological}. PRC satisfies this criterion by treating the mechanical body as a fixed, task-agnostic kernel, with only the output readout layer trained --- typically via linear regression~\cite{hauser2011towards}. 
Compared to traditional artificial neural networks (ANNs), the simplicity of treating the physical body as the computing kernel offers significantly lower computational cost, reduced memory and energy demands, and fast training --- enabling its deployment in computation-embodied autonomous systems~\cite{hauser2021physical}.

\begin{figure}[t!]
    \centering
    \includegraphics[scale=0.86]{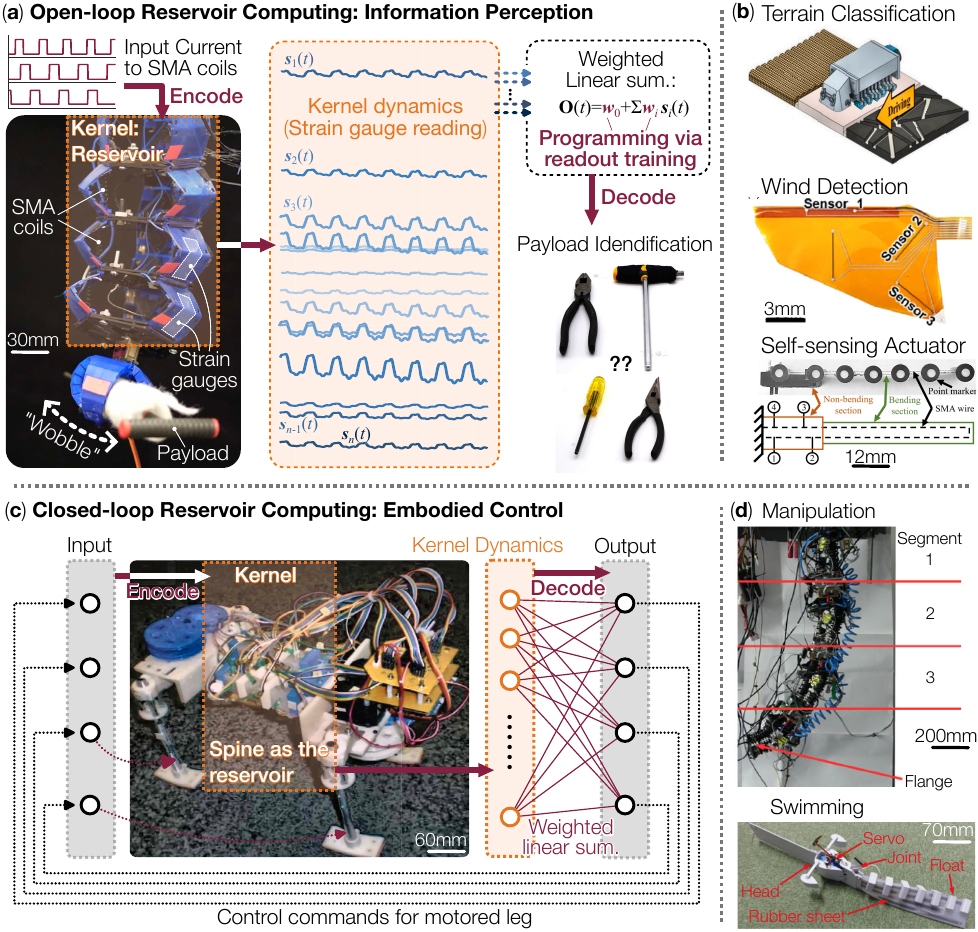}
    \caption{\textbf{Open and closed-loop physical reservoir computing in soft robots.}
    (a) Open-loop PRC for information perception: A modular manipulator with embedded strain gauges is driven by SMA actuators. Its high-dimensional body dynamics (measured by the strain $s_i(t)$) serve as reservoir states, which can be processed with trained linear readout $w_i$ to decode and identify the payload (image adapted from \cite{wang2025embodied} with permission). 
    (b) Other examples of information perception with open-loop PRC, including terrain classification(image adapted from \cite{tayama2025development} CC BY-ND 4.0), wind detection on a compliant membrane wing (\textcolor{black}{image adapted from \cite{tanaka2021flapping}, CC BY 4.0}), and a self-sensing shape memory alloy actuator that could predict its end effector trajectory (\textcolor{black}{image adapted from \cite{shougat2023self} with permission}).
    (c) Closed-loop PRC for embodied control: A quadruped robot uses its compliant spine as the reservoir.  The four outputs of the reservoir kernel are fed back to the leg actuators to generate robust and adaptable locomotion gaits (\textcolor{black}{image adapted from \cite{2013SpineReservoir} with permission}). 
    (d) Other examples of control embodiments with closed-loop PRC, including manipulation with a multi-segment continuum arm (\textcolor{black}{image adapted from \cite{eder2018morphological} with permission}) and a surface-swimming robot (\textcolor{black}{adapted from \cite{horii2021physical}, CC BY 4.0}). Note that all scale bars are approximate.}
    \label{fig:PRC}
\end{figure}

There are two rigorously developed frameworks for using physical reservoir computing kernels. They have been demonstrated in a damped-mass-spring network~\cite{hauser2011towards, hauser2012role}, and they can guide the use of PRC in soft robots.  The first framework is \textbf{open-loop}, in which the mechanical system acts as a fixed nonlinear kernel and only a static linear readout is trained to process the temporal data streams~\cite{hauser2011towards}. The second framework is \textbf{closed-loop}, in which the reservoir’s outputs are fed back to the actuators to shape future inputs, thus stabilizing or switching physical computing under simple static feedback~\cite{hauser2012role}. Building on these two frameworks, one can deploy the open-loop PRC into robots for \emph{information perception} --- i.e., extracting and decoding meaningful information from the high-dimensional body dynamics. On the other hand, the closed-loop PRC can be used for \emph{embodied control} --- i.e., routing the reservoir computing outputs as control commands to the actuators, thus producing and modulating rhythmic body motions. In the following two subsections, we detail the working principles and applications of these two frameworks.

\subsection*{Open-Loop Reservoir: Information Perception with Body Dynamics}

\textcolor{black}{\underline{An example of open-loop reservoir applied to soft robots}}: The open-loop framework allows PRC to enhance the perception of soft robots by transforming their bodies into multi-modal computing sensors. A compelling demonstration of this is the modular manipulator equipped with SMA coil actuators and simple strain gauges \cite{wang2025embodied} (Figure\ref{fig:PRC}a). The manipulator’s nonlinear body dynamics serve directly as the source for physical reservoir computing. More specifically: 

\begin{itemize}
    \item \textcolor{black}{\textit{Input encoding}: When the manipulator grasps and lifts different payloads, its SMA actuators generate pulsed forces to ``wobble'' the body slightly. }
    \item \textcolor{black}{\textit{Kernel}: In this robot, the soft body itself is the kernel (or reservoir). As the SMA wobbles the manipulator and its payload, the resulting body vibration, denoted as $s_i(t)$, is captured by the strain gauges. Such a vibrational response is rich and nonlinear, so its spatiotemporal feature contains information about the weight and orientation of the payloads.}
    \item \textcolor{black}{\textit{Output decoding}: By performing a simple and analog weighted linear summation of these strain gauge readings (O(t)=$w_0+\sum w_i s_i(t)$), the robot can directly estimate the payload weigth and orientation, thus classifying them.}
    \item \textcolor{black}{\textit{Re-programming}: The readout weights $w_i$ in the output layer are trained by regression methods, which can be adjusted according to the particular computing task at hand.}
\end{itemize}

This example clearly illustrates how the open-loop reservoir computing framework enables soft robots to conduct spatiotemporal filtering through their intrinsic deformation dynamics, allowing them to extract complex information without requiring dense sensor arrays or extensive digital processing. As a result, numerous studies have emerged utilizing different soft robotic platforms. For instance, a fabric-based soft manipulator can estimate joint bending angle and payload weight simultaneously using only a few distributed pressure sensors~\cite{wang2025proprioceptive}. Contact dynamics in a soft arm allow for tactile sensing and object property estimation without electronic skins~\cite{yoshimura2024research}. For environment monitoring, a brush-like flexible sensor encodes surface textures through passive contact~\cite{tayama2025development}(Fig.\ref{fig:PRC}b, up), while in aerial applications, a flapping-wing robot detects wind direction directly from wing deformation, eliminating the need for airflow sensors~\cite{tanaka2021flapping}(Fig.\ref{fig:PRC}b, middle). A SMA reservoir~\cite{shougat2023self} demonstrates the capability of predicting the future trajectory of its end effector under various driving signals (Fig. \ref {fig:PRC}b, bottom). Collectively, these demonstrations show that soft robots with sparse, low-dimensional sensors can nonetheless achieve high-dimensional perception by exploiting their own body as the physical reservoir---making PRC a minimalist yet powerful strategy for embodied sensing.

\subsection*{Closed-Loop Reservoir: Embodied Control}

\textcolor{black}{\underline{An example of closed-loop reservoir applied to soft robots}}: Beyond information perception, PRC enables soft robots to autonomously generate periodic and robust motions by embedding control into their intrinsic body dynamics. That is, instead of connecting robotic actuators to external digital controllers, one can feed the body reservoir's output back to these actuators for real-time motor behavior control. In this case, the deformation dynamics of the robotic body and its interaction with the environment play a critical role. A representative example is the quadruped robot with its flexible spine serving as the reservoir computing kernel \cite {2013SpineReservoir}(Fig.~\ref{fig:PRC}c). More specifically, 

\begin{itemize}

    \item \textcolor{black}{\textit{Kernel}: In this robot, its flexible spine is the kernel (or reservoir). Its intrinsic body dynamics are rich and nonlinear, capable of projecting input signals into a high-dimensional state vector.}
    
    \item \textcolor{black}{\textit{Input encoding and output decoding}: In this case, the linear readout layer performs weighted linear summations, mapping the internal force and strain of the flexible spine into four control commands, each for a motored leg. The readouts are first trained in an open-loop setup with teacher forcing. Once the training is complete, the loop is closed. As a result, the reservoir computer's output is also the input, eventually creating a self-sustained locomotion gait.} 
    
    \item \textcolor{black}{\textit{Re-programming}: Once the loop is closed, and the robot can perform trotting, bounding, or turning --- with a strong ability to recover from disturbance --- simply by switching the readout weights. This embodied controller demonstrates how PRC converts compliant mechanics into an energy-efficient control system: the body remains unchanged, only the readout is ``programmed,'' and feedback routes it from perception to action.}
    
\end{itemize}

This closed-loop, reservoir-enabled control principle has been successfully implemented across multiple platforms, including soft silicone arms~\cite{nakajima2014exploiting, eder2018morphological}(Fig.\ref{fig:PRC}d), tensegrity robots~\cite{caluwaerts2013locomotion}, and origami-inspired machines~\cite{bhovad2021physical}, all demonstrating motor primitives and robust dynamic behaviors. A pneumatic soft robotic arm, for instance, learns different end-effector trajectories and autonomously recovers from disturbances by exploiting its intrinsic dynamic richness~\cite{eder2018morphological}(Fig.\ref{fig:PRC}d, up). More recently, soft robots have also been shown to switch behaviors under varying environmental conditions by using reservoir systems that simultaneously encode control and sensory feedback~\cite{horii2021physical}(Fig.\ref{fig:PRC}d, bottom). Collectively, these demonstrations show that physical reservoir computing not only simplifies and stabilizes motion generation but also enables behavior switching through embodied computation, offering an energy-efficient alternative to conventional digital control architectures.

\textcolor{black}{\underline{Challenges and opportunities of analog physical reservoirs}: PRC offers an appealing framework to embody computation directly into soft robotic systems. In this framework, one can directly "multipurpose" a soft robotic body into a reservoir without substantial redesign, and quickly switch the computing function by adjusting the readout weights. However, reservoir computing also faces several fundamental challenges. The first is repeatability. Real-world physical systems can not guarantee identical dynamic output across multiple experiments; slight variations in fabrication, boundary conditions, temperature, and material behavior, along with drift and aging, can shift the reservoir's dynamic responses and degrade computing performance. A second challenge is noise, as real hardware inevitably introduces sensing and actuation noise that can be amplified by the reservoir's nonlinear dynamics, leading to degraded or even unstable output. The third challenge involves scaling. One can always increase the number of reservoir states to improve performance, but it means more sensors, more wires, and heavier data-processing burdens. Possible improvements across these areas include better operating-point stabilization and calibration procedures, noise-aware training with improved sensing electronics, and more efficient sensing architectures or dimensionality-reduction strategies that capture the essential physical dynamics without overwhelming the hardware. These efforts collectively point toward more reliable, robust, and scalable physical reservoir computing systems for soft robotics.}

\section{Algorithmic Physical Computing and Mechano-Logic}

Unlike analog computing, an algorithmic computer uses abstract logical rules to drive the input-output evolution. Correspondingly, their input and output signals are typically in a discrete format (e.g., binary 0-1, on-off, or true-false bits). Our omnipresent, CMOS-based digital computers are built almost exclusively on an algorithmic architecture, relying on binary data streams passing through nested Boolean logic gates to perform computations~\cite{jaeger2023toward}. 
However, one can also achieve algorithmic computing without electronics \cite{yasuda2021mechanical}. That is, instead of electrons flowing through binary logic gates, one can construct binary components that operates with elastic deformations, fluid flows, or other physical stimuli. Each physical component can act as an equivalent to logic gates, memory cells, or timing elements to fulfill computation roles. Though fundamentally different in shape and format, the underlying goals of digital and physical algorithmic computers remain the same: to perform computation tasks by following programmed logical rules for information processing.

\subsection*{Mechanical Bistable Mechanisms as the Building Blocks}

It is worth highlighting the important role of bistable mechanisms in physical algorithmic computing because they can directly emulate the 0-1 binary states of CMOS electronics. Bi-stability --- defined as a physical construct’s ability to settle into two distant stable equilibria without additional external aids --- arises from material or geometric nonlinearities and can be implemented using curved elastic beams~\cite{chen2018harnessing, raney2016stable}, elastomeric membranes~\cite{rothemund2018soft, patel2023highly}, or origami folds~\cite{huang2022bistable, kaufmann2022harnessing, chen2025advances}. 
A bistable mechanism naturally exhibits large and rapid deformation when it snaps between its stable equilibria~\cite{pal2021exploiting}, thereby amplifying the actuation output and simplifying the control of the soft robot~\cite{tang2020leveraging}. For example, snapping elastic caps convert slow inflation into explosive jumping~\cite{gorissen2020inflatable}, bistable curved fins enable fast, high-efficiency swimming~\cite{chi2022snapping}. They can also help program the robotic deformation, like in the soft sheets with an array of snap-through domes~\cite{faber2020dome}. 
A bistable mechanism can also perform sensing and, therefore, encode inputs into the physical computer. For instance, a skin-like sensing surface with localized snap-through cells can act as a mechanical signal amplifiers that translate pressure or contact into discrete mechanical states~\cite{le2019filtered}. Soft mechanosensors based on bistable structures can provide binary contact information without continuous electrical feedback~\cite{thuruthel2020bistable}

Most importantly (and most relevant to this paper), the bistable mechanism can serve as a mechanical analog to transistors, functioning as one-bit memory units by switching between two stable configurations. These configurations can be mapped to binary states `0' and `1' based on the input force, pressure, or displacement, enabling the construction of logic gates and sequential logic circuits using entirely mechanical components~\cite{ramachandran2016elastic, holmes2013control, yang2010thermopneumatically, maffli2013mm, yasuda2021mechanical}. Therefore, mechanical bi-stability provides a fundamental means to encode, store, and process information within a robot’s physical body. Typically, these robotic algorithmic computers directly borrow the design and architecture of CMOS-based systems, but they can be inherently energy efficient, retaining their state without continuous power input.

\subsection*{Examples of Embodied Algorithmic Computing}
\begin{figure}[t!]
    \centering
    \includegraphics[scale=0.95]{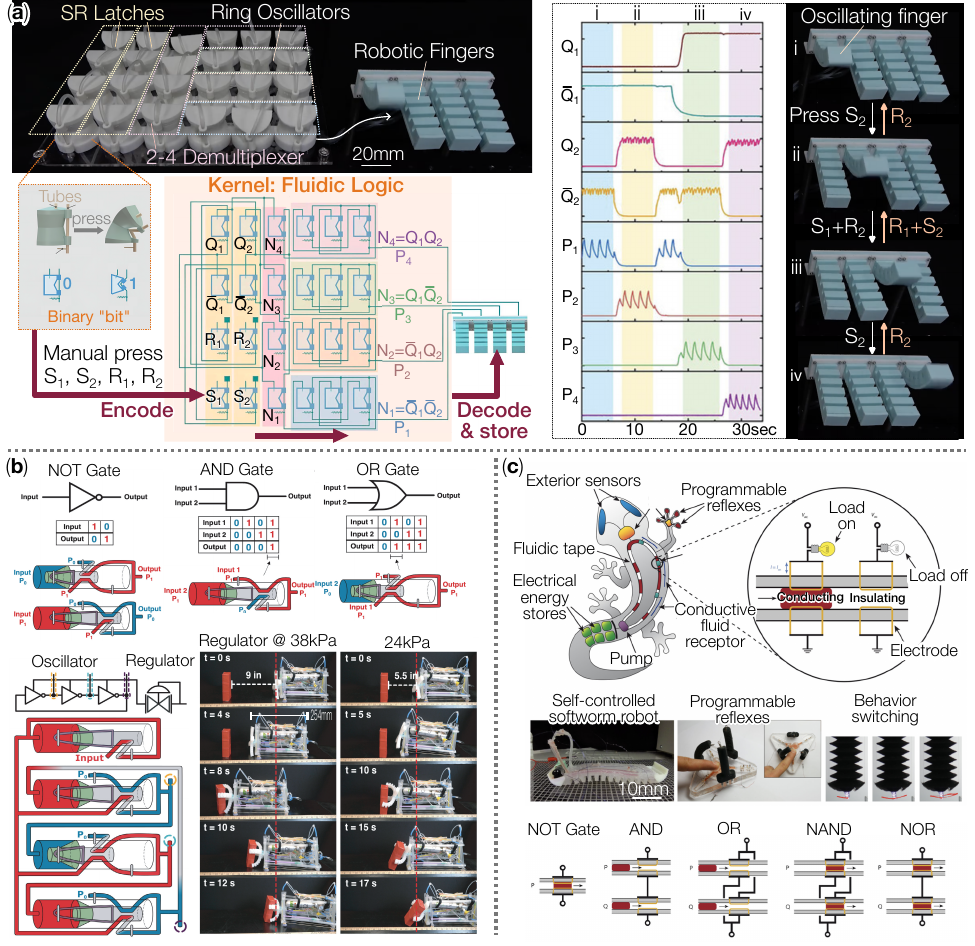}
    \caption{\textbf{Physical Algorithmic Computing}. (a) Reprogrammable metamaterial processors with robotic fingers: Fluidic unit-cells with 0-1 binary states are connected to create mechanical logic circuitry to control finger actions  (\textcolor{black}{adapted from \cite{jiao2024reprogrammable}, CC BY 4.0}). (b) Complementary soft pneumatic valves: Piston-based, four-terminal modules are paired to achieve Boolean logic operation, non-volatile latches, and analog pressure regulation. Then they are integrated with sub-circuits to create ring oscillators and counters to control crawling robots and wearable devices (\textcolor{black}{adapted from \cite{decker2022programmable}, CC BY-NC-ND 4.0}). (c) Soft-matter computer: Conductive-fluid receptors transduce spatiotemporal fluid patterns into electrical drives, which can realize analog filtering, amplification, and logic gates with simple composition. As a result, such conductive fluidic mechanism enables on-body control for locomotion, reflexive grasping, and behavior switching (image adapted from \cite{garrad2019soft} with permission). Note that all scale bars are approximate.}
    \label{fig:PAC}
\end{figure}

To better illustrate the working principle, we detail a fluidic and algorithmic computer based on a reprogrammable metamaterial processor (Fig. \ref{fig:PAC}a). The processor comprises identical bistable unit cells whose elastomeric chambers snap at defined pressure thresholds, converting vacuum and atmospheric pressure input into binary states {\cite{jiao2024reprogrammable}}. As a result, a bistable unit cell with a clever tubing design can function like a resistor, enabling the construction of complex logical circuitry. More specifically

\begin{itemize}
    \item \textcolor{black}{\textit{Kernel}: In the example, the kernel has 24 unit cells that are connected into a soft processor, including two SR latches, a 2–4 demultiplexer, and four ring oscillators (each linked to a soft robotic finger). The soft processor and the fingers are all powered by one constant vacuum pressure. The entire system can reversibly switch between four different operation modes, each of which corresponds to the oscillatory bending of one finger.}
    
    \item \textcolor{black}{\textit{Input encoding}: The operator can choose the operating mode via manually pressing the input cells of the SR latches (\textit{input encoding}). For example, if the first figure is activated initially, one can press the `$S_2$' cell to activate the second figure and then press the `$R_2$' cell to switch back. The outputs of the two SR latches are sent to the demultiplexer, so these 2 data lines are converted into 4. The resulting four outputs of the demultiplexer serve as the power source for the ring oscillators.}

    \item \textcolor{black}{\textit{Output decoding}: The robotic fingers transform the output oscillatory pressures into mechanical bending. Notably, the current operation mode persists even after the removal of the pressing force, owing to the ability of the SR latches to retain their logic states until updated by new inputs.}

    \item \textcolor{black}{\textit{Re-programming}: Finally, one can re-arrange these fluidic unit cells to construct a new soft processor with different input-output mapping.}
\end{itemize}

In addition to the example above, there are several other attempts to copy the electronic logic circuitry into the fluidic domain and construct fluid Boolean gates with soft complementary valves, ring oscillators, and modular cells \textcolor{black}{\cite{song2021cmos, decker2022programmable, tracz2022tube,conrad20243d, liu2023discriminative,mahon2019soft,stanley2024high} }(Fig. \ref{fig:PAC}b).
Besides the pressurized fluidics, algorithmic computing can also be implemented with other novel materials and multi-stable mechanisms. Here, we list four additional approaches: 1) \emph{Conductive fluidics}: Conductive fluidic receptors (CFRs) embedded in soft structures can act as hybrid mechanical-electrical logic units, enabling soft matter computers to perform sensing, logic, and actuation all in one continuous system {\cite{bartlett20153d, garrad2019soft,yue2025embodying}} (Fig. \ref{fig:PAC}c). 2) \emph{Magnetic fluidics}: Magnetic liquid metal droplets can create flexible and reconfigurable logic gates with decoupled input/output channels and multi-modal control using phase-state transitions {\cite{xu2023soft}}. 3) \emph{Heat responsive materials}: Mechanical logic has also been achieved using mechanical and multiplexed switches that integrate bistable beams and thermally responsive materials to perform logic operations and mechanical memory storage {\cite{yan2023origami, jiao2024reprogrammable,li2021light,yang2025review}}. 4) \emph{Multi-stable mechanisms}: Finally, algorithmic computing is also possible with pure elastic force and deformation. Unique architectures, such as counter-snapping metamaterials, provide logic behavior via geometric nonlinearity, where structural instability enables programmable stiffness transitions and collective switching sequences, making them useful for timing and computation {\cite{ducarme2025exotic}}. Recent advances in modular chiral origami metamaterials further expand this logic repertoire by introducing multi-stable and reprogrammable architectures that can store information through mechanically encoded hysteresis and non-commutative state transitions {\cite{zhao2025modular}}.

It is worth noting that physical algorithmic computing can also enable locomotion generation and sequencing --- locomotion turns out to be the robot task shared by all physical computers reviewed in this study. In this regard, algorithmic computing supports locomotion sequencing through timing control and built-in periodicity. For example, pneumatic ring oscillators and fluidic valve networks, constructed from bistable logic gates, have been used to generate self-sustained actuation cycles for crawling and walking gaits in soft quadrupeds and hexapods {\cite{decker2022programmable, liu2023discriminative, conrad20243d}}. Morphologically encoded logic and routing delays, utilizing internal resistance gradients, have also been employed to produce pressure wave propagation and staggered motion, enabling gait generation through a single input channel {\cite{matia2023harnessing}}. Reconfigurable metamaterials and origami systems offer structural ways to embed sequencing. For instance, modular soft metamaterial robots have been programmed to switch between gaits—turning, serpentine, reciprocating—by physically re-arranging submodules acting as logic units {\cite{jiao2024reprogrammable, liu2023discriminative}}. In another case, origami robots with memory registers and rotating read-heads perform controlled motion paths by storing finite-state instructions mechanically {\cite{yan2023origami}}.

\textcolor{black}{\underline{Challenges and opportunities of physical algorithmic computing}: Compared to analog physical computers, algorithmic physical computing is quite versatile in that it can borrow many designs and working principles from well-established CMOS electronics. However, the reviewed examples above lag behind in terms of speed and scaling. Their computing speed is constrained by the relatively slow physical processes, such as pressurizing and venting of fluidic networks, deformation of thick elastomeric chambers, and, in some cases, heat diffusion through responsive materials. Using physical signals instead of electric ones also makes miniaturization more challenging: The finite size of multi-stable unit cells, the need for compliant interconnection devices, and the risk of mechanical crosstalk between unit cells make routing and isolation harder than in CMOS. Therefore, significant research efforts are necessary. For example, using advanced manufacturing technology can help minimize the unit cell size and enable more integrated packaging, thus speeding up physical computers (as we discuss later in Section 5). Regardless, physical algorithmic computing is still a desirable choice for small-scale logic and simple on-board control (as we discuss further in the conclusion section). } 

\section{Perspective for Future Advancement}

\begin{figure}[t!]
    \centering
    \includegraphics[scale=0.85]{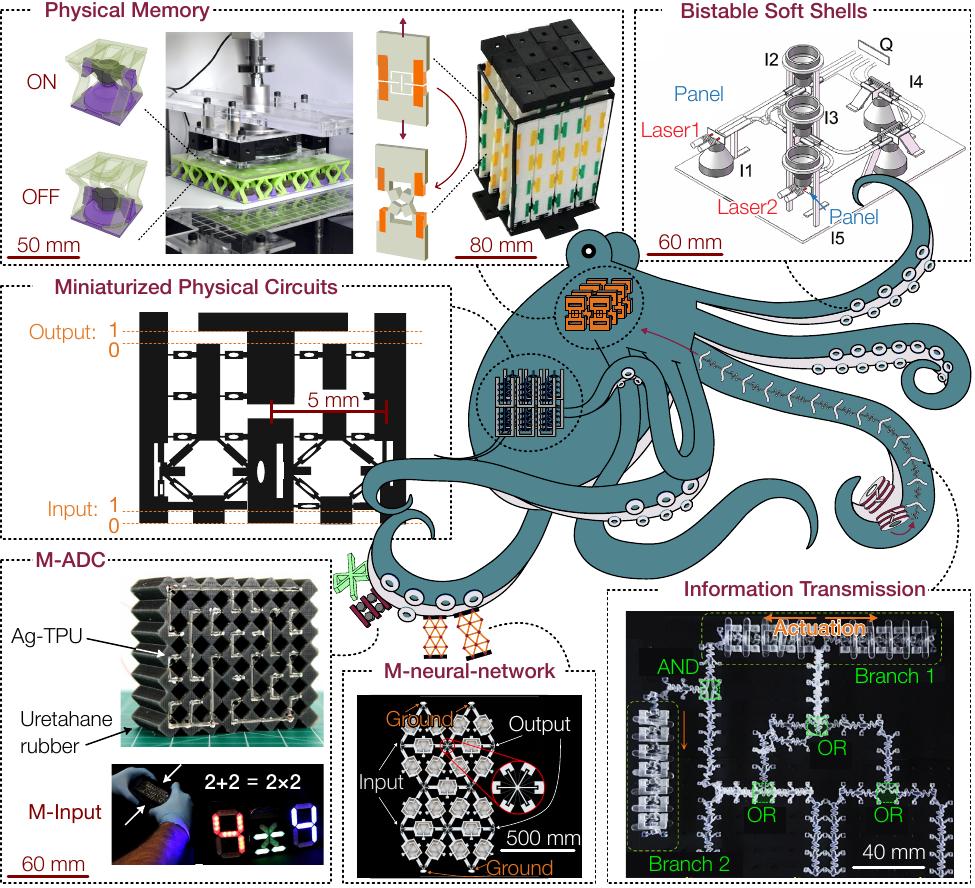}
    \caption{
    \textbf{Future directions for embodying physical computing in soft robots}: One can continue to advance the computing kernel's density and capacity in soft robots by adopting new strategies to process encoded inputs, either locally or through a centralized kernel. To illustrate some of the promising concepts, we envision an octopus-inspired soft robot with computational capabilities distributed across its tentacles as well as in its brain, following the mechanical computing framework described in this manuscript. Each component can encode, process, and decode data (with mechanical memory reserved for storage). Starting from the top right tentacle and moving clockwise: Bistable soft shells enable rule-changeable logic operations (image adapted from \cite{yang2025bistable} CC BY 4.0); Information processing during transmission via non-dispersive mechanical solitary waves (image adapted from \cite{byun2024integrated} CC BY 4.0); A mechanical neural network offloads computation and can be attached to the robot's skin (image adapted from \cite{hopkins2023using} with permission); Mechanical analog-to-digital converters can be embedded in the tentacles (image adapted from \cite{el2022mechanical} with permission). In the robot's brain, miniaturized physical circuits mimic an algorithmic logic unit (ALU) (image adapted from \cite{song2019additively} CC BY 4.0). Finally, reprogrammable and non-volatile mechanical memories can store data either with magnetic (left, image adapted from \cite{chen2021reprogrammable} with permission) or thermal principles (right, image adapted from \cite{meng2023encoding} CC BY 4.0). Note that all scale bars are approximate.
    }
    \label{fig:future}
\end{figure}

Since this perspective lies at the intersection of physical computing and soft robots, it is intuitive to ask questions about the future direction using the ``supply-and-demand'' analogy. On the supply side: ``Are there any newly available capabilities in physical computing that can be deployed for soft robotics?'' On demand side: ``What additional computing power would future soft robots require?'' \textcolor{black}{Here, our unique perspective of dissecting soft robots into the encoding, kernel, and decoding layers can offer a systematic framework to introduce new computing concepts. For example, one can keep the encoding (e.g., sensor and input) and decoding components (e.g., actuator) the same, and ``swap'' the kernel with different designs that have new kinds of computing capacity. Alternatively, one can ``upgrade'' the kernel with a more advanced design.} By surveying the current physical computing studies, one can discover many unique approaches that could be integrated into soft robots in the future (examples in Fig. \ref{fig:future}). 

\textcolor{black}{Here, we highlight three most promising topics from three different angles: \textit{function}, \textit{scale}, and \textit{system integration}. Regarding the function, the current physical computing has demonstrated an impressive ability to extract information from sensory signals and execute actuation commands. On the other hand, on-board non-volatile memory, a vital component in modern computing paradigms, has yet to be implemented in soft robots. Regarding scale, the current physical computing in soft robots is relatively large in terms of physical size. Minimizing their scale using high-precision manufacturing techniques might help advance the performance and reliability of physical computers to a new level, addressing some of the challenges in physical computing as we discussed earlier. Finally, regarding systematic integration, the current physical computing setup in soft robots primarily operates in a standalone manner. However, some integration with digital hardware (e.g., for long-distance communication) can enhance the overall capability. Therefore, a meaningful and integrated mechanical-electrical hybrid circuit can be advantageous. In the following section, we present recent studies in these three aspects.}

\subsection*{Physical Memory} 
An advanced and autonomous robot should also be able to memorize the operator's instructions or the knowledge from its interaction with the working environment. In this regard, we have seen some promising examples of memory in the mechanical metamaterial domain. That is, by combining responsive materials (memory encoding/decoding) and elastic bistability (storage), mechanical metamaterials can achieve information storage via mechanical bits (m-bits), similar to their digital counterparts. These m-bits can be \textcolor{black}{bistable elastic shell to realize mechano-fluidic memory\cite{comoretto2025embodying}, or} bistable origami structures \cite{yasuda2017origami} or tiles of bistable Kirigami units \cite{xin2024magnetic}, forming 2D and 3D storage arrays. Examples include temperature-responsive bistable Kirigami units \cite{meng2023encoding, yang2025role}, which sequentially retrieve stored information in 3D arrays, and magnetic-responsive bistable elastic shell units \cite{chen2021reprogrammable}, which enable on-demand re-programmability for 2D arrays. These configurations function as non-volatile mechanical memories and could be used in soft robotics in the future. A key challenge will be designing an integrative approach to encode information from the physical memory into the computing kernel, and subsequently decode and store the computing output into the physical memory device. 

\subsection*{Miniaturized and Higher-Capacity Physical Circuitry}
Just as CMOS-based computers' never-ending quest to shrink the size of their basic electronic units, soft robots can also benefit from a smaller and more capable physical computer onboard. Early physical algorithmic computers --- such as the waterbomb origami with bistable hinges \cite{meng2021bistability} --- were bulky and limited to AND, OR, and NOT gates. More compact designs like bistable curved beam arrays have introduced NOR and NAND operations \cite{mei2021mechanical}. Bistable soft shells allowed re-programmable mechanologics so that their operation mode can be switched on demand (e.g., from XOR to XNOR)\cite{yang2025bistable}. Additionally, self-powered origami mechanologics \cite{zhang2023meta} and thermal mechanical transistors \cite{chen2024thermal} completed the binary logic set with XNOR and XOR gates. More recently, with advancements in sub-millimeter additive manufacturing, small-scale mechanologics based on buckling micro-flexures have emerged \cite{song2019additively, yue2024flexibly}. Integrating these mechanologics has led to fully mechanical half-adders \cite{yue2024flexibly, mei2021mechanical, yang2025bistable, zhang2023meta}, full-adders \cite{yang2025bistable, zhang2023meta, chen2024thermal}, and solitary wave-based mechanical computing platforms \cite{byun2024integrated}. It is not hard to imagine that some of these miniaturized and physical algorithmic units will be integrated into soft robots in the future, enabling fully onboard computation and control.  

\subsection*{Integrated, Mechanical-Electrical Hybrid Circuits}
While physical computing aspires to perform computation tasks without complex CMOS-based electronics, it can still benefit from using some simple electronic components. Indeed, the physical reservoir computer reviewed in this study is a mechanical-electric hybrid system, because its readout layer requires electronics to perform weighted linear summation (e.g., using an analog adder circuit with an Op-Amp). This provides the physical reservoir with excellent re-programmability and multi-tasking abilities that are not yet available in analog oscillators or physical algorithmic computers. 

Therefore, mechanical-electrical hybrid circuits could provide scalable computing capabilities for future soft robots. These hybrid systems convert mechanical input into electrical signals by opening or closing conductive pathways in response to deformation. Applications include mechanical digital sensors \cite{nick2020liquid}, mechanical analog-to-digital converters (m-ADC) \cite{hyatt2023programming}, and mechanical arithmetic logic units (m-ALU), which embed Boolean logic into soft configurable structures \cite{el2022mechanical, el2021digital, xi2025kinematically}. Higher-level computation becomes feasible when memory is integrated into computation, as seen in in-memory mechanical computing platforms, such as mechanical neural networks \cite{mei2023memory} and linear equation solvers \cite{el2024intelligent}. These hybrid circuits have yet to be integrated into soft robotics, but they present an exciting pathway for systematic integrations.  


\section{Conclusion: A Robust Route to Mechanical Intelligence}
In summary, the convergence of physical computing with soft robotics is a promising strategy for softening and onboarding control.  By integrating a physical computing kernel --- such as an analog oscillator, a physical reservoir computer, or an algorithmic computer --- into a soft robotic system, these robots can achieve sophisticated locomotion and manipulation tasks that would typically require a conventional digital control.  

For example, this perspective paper demonstrates that an electronics-free legged robot equipped with an analog oscillator can perform coordinated locomotion and reverse direction upon encountering an obstacle. The soft modular manipulator, with inherent physical reservoir computing capacity, can utilize its body dynamics to estimate the weight and orientation of its payload, enabling it to classify the payload without relying on electronic sensors, such as cameras.  Another soft robotic hand integrated with an algorithmic fluidic circuitry that can operate based on abstract Boolean logical rules. 

\textcolor{black}{However, it is worth noting that all of the physically computing robots in this perspective remain at the proof-of-concept level. Implementing these exciting concepts into practical, real-world use still requires a significant amount of research efforts and system engineering.}

\textcolor{black}{Despite the rapid advances in this field, it is unlikely that physical computers embedded in soft robots can catch up with digital hardware in terms of computational speed and information density in the foreseeable future. Therefore, it is unrealistic to replace conventional digital hardware entirely with physical computing. Instead, engineers must answer the critical question: "How much should we use physical computing?" "Where to apply them?" and "How can we seamlessly integrate physical computing with conventional digital hardware?" }

\textcolor{black}{For robots, physical computing is advantageous because of its softness, simplicity, and robustness. Therefore, it makes the most sense to use physical computing in the following three scenarios. 1) The targeted tasks are closely related to the robot's physical body --- this is why we have seen great success in locomotion generation and information extraction via direct physical interaction (using physical reservoir computing). 2) The robots need to be small and entirely soft --- because physical computing could seamlessly integrate with the soft robotic body without the complexity of adding electronic components (e.g., using advanced 3D printing). 3) The working conditions are demanding --- for example, fluidics-based computation is desirable for underwater operations, where electronics are vulnerable to damage. On the other hand, conventional digital electronics is more suitable for ``over the distance'' tasks, such as obstacle avoidance using vision data or long-distance communication with operators. }

\textcolor{black}{Therefore, the future of physically computing robots hinges upon two pillars: the continual advances in physical computing and its strategic integration with conventional digital hardware. As we discussed in Section 5, in the foreseeable future, we are likely to witness the creation of more powerful physical computing thanks to miniaturization and integrated memory capacity. With a more advanced physical computing kernel, a soft robot can acquire information from interacting with the surrounding environment, memorize the acquired knowledge, and execute the action plan, all in a highly integrated mechanical domain. On the other hand, new strategies will emerge to tightly integrate physical computing with digital computing via novel mechanical-electrical hybrid circuits, enabling physical computing robots to operate within large-scale automated systems. This vision of soft robots hinges on the ongoing convergence of various engineering disciplines, including mechanical metamaterials, computing theory, advanced manufacturing, and interdisciplinary design. }

\section*{Acknowledgments}
The authors acknowledge the support from the National Science Foundation (CMMI-2312422, 2328522, EFRI-2422340) and Virginia Tech (via Startup Fund and Graduate Student Assistantship).

\bibliography{scibib}

@article{meng2023encoding,
  title={Encoding and storage of information in mechanical metamaterials},
  author={Meng, Zhiqiang and Yan, Hujie and Liu, Mingchao and Qin, Wenkai and Genin, Guy M and Chen, Chang Qing},
  journal={Advanced Science},
  volume={10},
  number={20},
  pages={2301581},
  year={2023},
  publisher={Wiley Online Library}
}

@article{hopkins2023using,
  title={Using binary-stiffness beams within mechanical neural-network metamaterials to learn},
  author={Hopkins, Jonathan B and Lee, Ryan H and Sainaghi, Pietro},
  journal={Smart Materials and Structures},
  volume={32},
  number={3},
  pages={035015},
  year={2023},
  publisher={IOP Publishing}
}

@article{el2022mechanical,
  title={Mechanical integrated circuit materials},
  author={El Helou, Charles and Grossmann, Benjamin and Tabor, Christopher E and Buskohl, Philip R and Harne, Ryan L},
  journal={Nature},
  volume={608},
  pages={699--703},
  year={2022},
  publisher={Nature Publishing Group UK London}
}

@article{chen2021reprogrammable,
  title={A reprogrammable mechanical metamaterial with stable memory},
  author={Chen, Tian and Pauly, Mark and Reis, Pedro M},
  journal={Nature},
  volume={589},
  number={7842},
  pages={386--390},
  year={2021},
  publisher={Nature Publishing Group UK London}
}

@article{yasuda2017origami,
  title={Origami-based tunable truss structures for non-volatile mechanical memory operation},
  author={Yasuda, Hiromi and Tachi, Tomohiro and Lee, Mia and Yang, Jinkyu},
  journal={Nature communications},
  volume={8},
  number={1},
  pages={962},
  year={2017},
  publisher={Nature Publishing Group UK London}
}

@article{meng2021bistability,
  title={Bistability-based foldable origami mechanical logic gates},
  author={Meng, Zhiqiang and Chen, Weitong and Mei, Tie and Lai, Yuchen and Li, Yixiao and Chen, CQ},
  journal={Extreme Mechanics Letters},
  volume={43},
  pages={101180},
  year={2021},
  publisher={Elsevier}
}

@article{mei2023memory,
  title={In-memory mechanical computing},
  author={Mei, Tie and Chen, Chang Qing},
  journal={Nature Communications},
  volume={14},
  number={1},
  pages={5204},
  year={2023},
  publisher={Nature Publishing Group UK London}
}

@article{yasuda2021mechanical,
  title={Mechanical computing},
  author={Yasuda, Hiromi and Buskohl, Philip R and Gillman, Andrew and Murphey, Todd D and Stepney, Susan and Vaia, Richard A and Raney, Jordan R},
  journal={Nature},
  volume={598},
  number={7879},
  pages={39--48},
  year={2021},
  publisher={Nature Publishing Group UK London}
}

@article{zhang2021wirelessly,
  title={Wirelessly actuated thermo-and magneto-responsive soft bimorph materials with programmable shape-morphing},
  author={Zhang, Jiachen and Guo, Yubing and Hu, Wenqi and Sitti, Metin},
  journal={Advanced Materials},
  volume={33},
  number={30},
  pages={2100336},
  year={2021},
  publisher={Wiley Online Library}
}

@article{fan2020reconfigurable,
  title={Reconfigurable multifunctional ferrofluid droplet robots},
  author={Fan, Xinjian and Dong, Xiaoguang and Karacakol, Alp C and Xie, Hui and Sitti, Metin},
  journal={Proceedings of the National Academy of Sciences},
  volume={117},
  number={45},
  pages={27916--27926},
  year={2020},
  publisher={National Academy of Sciences}
}

@article{pal2021exploiting,
  title={Exploiting mechanical instabilities in soft robotics: control, sensing, and actuation},
  author={Pal, Aniket and Restrepo, Vanessa and Goswami, Debkalpa and Martinez, Ramses V},
  journal={Advanced Materials},
  volume={33},
  number={19},
  pages={2006939},
  year={2021},
  publisher={Wiley Online Library}
}

@article{goswami20193d,
  title={3D-architected soft machines with topologically encoded motion},
  author={Goswami, Debkalpa and Liu, Shuai and Pal, Aniket and Silva, Lucas G and Martinez, Ramses V},
  journal={Advanced functional materials},
  volume={29},
  number={24},
  pages={1808713},
  year={2019},
  publisher={Wiley Online Library}
}

@article{zhao2022twisting,
  title={Twisting for soft intelligent autonomous robot in unstructured environments},
  author={Zhao, Yao and Chi, Yinding and Hong, Yaoye and Li, Yanbin and Yang, Shu and Yin, Jie},
  journal={Proceedings of the National Academy of Sciences},
  volume={119},
  number={22},
  pages={e2200265119},
  year={2022},
  publisher={National Academy of Sciences}
}

@article{yang202088,
  title={An 88-milligram insect-scale autonomous crawling robot driven by a catalytic artificial muscle},
  author={Yang, Xiufeng and Chang, Longlong and P{\'e}rez-Arancibia, N{\'e}stor O},
  journal={Science Robotics},
  volume={5},
  number={45},
  pages={eaba0015},
  year={2020},
  publisher={American Association for the Advancement of Science}
}

@article{zhou2024self,
  title={Self-Sustained and Coordinated Rhythmic Deformations with SMA for Controller-Free Locomotion},
  author={Zhou, Ziyang and Li, Suyi},
  journal={Advanced Intelligent Systems},
  volume={6},
  number={5},
  pages={2300667},
  year={2024},
  publisher={Wiley Online Library}
}

@article{kotikian2019untethered,
  title={Untethered soft robotic matter with passive control of shape morphing and propulsion},
  author={Kotikian, Arda and McMahan, Connor and Davidson, Emily C and Muhammad, Jalilah M and Weeks, Robert D and Daraio, Chiara and Lewis, Jennifer A},
  journal={Science robotics},
  volume={4},
  number={33},
  pages={eaax7044},
  year={2019},
  publisher={American Association for the Advancement of Science}
}

@article{dickinson2000animals,
  title={How animals move: an integrative view},
  author={Dickinson, Michael H and Farley, Claire T and Full, Robert J and Koehl, MAR and Kram, Rodger and Lehman, Steven},
  journal={science},
  volume={288},
  number={5463},
  pages={100--106},
  year={2000},
  publisher={American Association for the Advancement of Science}
}

@article{zhou2023cpg,
  title={A CPG-Based Versatile Control Framework for Metameric Earthworm-Like Robotic Locomotion},
  author={Zhou, Qinyan and Xu, Jian and Fang, Hongbin},
  journal={Advanced Science},
  volume={10},
  number={14},
  pages={2206336},
  year={2023},
  publisher={Wiley Online Library}
}

@article{drotman2021electronics,
  title={Electronics-free pneumatic circuits for controlling soft-legged robots},
  author={Drotman, Dylan and Jadhav, Saurabh and Sharp, David and Chan, Christian and Tolley, Michael T},
  journal={Science Robotics},
  volume={6},
  number={51},
  pages={eaay2627},
  year={2021},
  publisher={American Association for the Advancement of Science}
}

@article{horsman2014does,
  title={When does a physical system compute?},
  author={Horsman, Dominic and Stepney, Susan and Wagner, Rob C and Kendon, Viv},
  journal={Proceedings of the Royal Society A: Mathematical, Physical and Engineering Sciences},
  volume={470},
  number={2169},
  pages={20140182},
  year={2014},
  publisher={The Royal Society Publishing}
}

@article{he2023modular,
  title={A modular strategy for distributed, embodied control of electronics-free soft robots},
  author={He, Qiguang and Yin, Rui and Hua, Yucong and Jiao, Weijian and Mo, Chengyang and Shu, Hang and Raney, Jordan R},
  journal={Science advances},
  volume={9},
  number={27},
  pages={eade9247},
  year={2023},
  publisher={American Association for the Advancement of Science}
}

@article{wang2023versatile,
  title={A versatile jellyfish-like robotic platform for effective underwater propulsion and manipulation},
  author={Wang, Tianlu and Joo, Hyeong-Joon and Song, Shanyuan and Hu, Wenqi and Keplinger, Christoph and Sitti, Metin},
  journal={Science Advances},
  volume={9},
  number={15},
  pages={eadg0292},
  year={2023},
  publisher={American Association for the Advancement of Science}
}

@article{wang20233d,
  title={3D-Printed Photoresponsive Liquid Crystal Elastomer Composites for Free-Form Actuation},
  author={Wang, Yuchen and Yin, Rui and Jin, Lishuai and Liu, Mingzhu and Gao, Yuchong and Raney, Jordan and Yang, Shu},
  journal={Advanced Functional Materials},
  volume={33},
  number={4},
  pages={2210614},
  year={2023},
  publisher={Wiley Online Library}
}

@article{mu2023spiral,
  title={Spiral-Shape Fast-Moving Soft Robots},
  author={Mu, Weilei and Li, Mengjiao and Chen, Erdong and Yang, Yiduo and Yin, Jie and Tao, Xiaoming and Liu, Guijie and Yin, Rong},
  journal={Advanced Functional Materials},
  volume={33},
  number={35},
  pages={2300516},
  year={2023},
  publisher={Wiley Online Library}
}

@article{rus2015design,
  title={Design, fabrication and control of soft robots},
  author={Rus, Daniela and Tolley, Michael T},
  journal={Nature},
  volume={521},
  number={7553},
  pages={467--475},
  year={2015},
  publisher={Nature Publishing Group UK London}
}

@article{boyvat2021remote,
  title={Remote modular electronics for wireless magnetic devices},
  author={Boyvat, Mustafa and Sitti, Metin},
  journal={Advanced Science},
  volume={8},
  number={17},
  pages={2101198},
  year={2021},
  publisher={Wiley Online Library}
}

@article{ke2024synergistical,
  title={Synergistical mechanical design and function integration for insect-scale on-demand configurable multifunctional soft magnetic robots},
  author={Ke, Xingxing and Yong, Haochen and Xu, Fukang and Chai, Zhiping and Jiang, Jiajun and Ni, Xiang and Wu, Zhigang},
  journal={Soft Robotics},
  volume={11},
  number={1},
  pages={43--56},
  year={2024},
  publisher={Mary Ann Liebert, Inc., publishers 140 Huguenot Street, 3rd Floor New~…}
}

@article{soon2023pangolin,
  title={Pangolin-inspired untethered magnetic robot for on-demand biomedical heating applications},
  author={Soon, Ren Hao and Yin, Zhen and Dogan, Metin Alp and Dogan, Nihal Olcay and Tiryaki, Mehmet Efe and Karacakol, Alp Can and Aydin, Asli and Esmaeili-Dokht, Pouria and Sitti, Metin},
  journal={Nature Communications},
  volume={14},
  number={1},
  pages={3320},
  year={2023},
  publisher={Nature Publishing Group UK London}
}

@article{luo2023autonomous,
  title={Autonomous self-burying seed carriers for aerial seeding},
  author={Luo, Danli and Maheshwari, Aditi and Danielescu, Andreea and Li, Jiaji and Yang, Yue and Tao, Ye and Sun, Lingyun and Patel, Dinesh K and Wang, Guanyun and Yang, Shu and others},
  journal={Nature},
  volume={614},
  number={7948},
  pages={463--470},
  year={2023},
  publisher={Nature Publishing Group UK London}
}

@article{wang2022adaptive,
  title={Adaptive wireless millirobotic locomotion into distal vasculature},
  author={Wang, Tianlu and Ugurlu, Halim and Yan, Yingbo and Li, Mingtong and Li, Meng and Wild, Anna-Maria and Yildiz, Erdost and Schneider, Martina and Sheehan, Devin and Hu, Wenqi and others},
  journal={Nature communications},
  volume={13},
  number={1},
  pages={4465},
  year={2022},
  publisher={Nature Publishing Group UK London}
}

@article{wu2023caterpillar,
  title={Caterpillar-inspired soft crawling robot with distributed programmable thermal actuation},
  author={Wu, Shuang and Hong, Yaoye and Zhao, Yao and Yin, Jie and Zhu, Yong},
  journal={Science Advances},
  volume={9},
  number={12},
  pages={eadf8014},
  year={2023},
  publisher={American Association for the Advancement of Science}
}

@article{he2024modular,
  title={Modular Stimuli-Responsive Valves for Pneumatic Soft Robots},
  author={He, Qiguang and Yin, Rui and Hua, Yucong and Shu, Hang and Zhu, Xiaoheng and Haque, ABM Tahidul and Ferracin, Samuele and Patel, Saheli and Jiao, Weijian and Raney, Jordan R},
  journal={Advanced Intelligent Systems},
  pages={2400659},
  year={2024},
  publisher={Wiley Online Library}
}

@article{xin2024magnetic,
  title={Magnetic Poles Enabled Kirigami Meta-Structure for High-Efficiency Mechanical Memory Storage},
  author={Xin, Libiao and Li, Yanbin and Wang, Baolong and Li, Zhiqiang},
  journal={Advanced Functional Materials},
  volume={34},
  number={9},
  pages={2310969},
  year={2024},
  publisher={Wiley Online Library}
}

@article{song2019additively,
  title={Additively manufacturable micro-mechanical logic gates},
  author={Song, Yuanping and Panas, Robert M and Chizari, Samira and Shaw, Lucas A and Jackson, Julie A and Hopkins, Jonathan B and Pascall, Andrew J},
  journal={Nature communications},
  volume={10},
  number={1},
  pages={882},
  year={2019},
  publisher={Nature Publishing Group UK London}
}

@article{mei2021mechanical,
  title={A mechanical metamaterial with reprogrammable logical functions},
  author={Mei, Tie and Meng, Zhiqiang and Zhao, Kejie and Chen, Chang Qing},
  journal={Nature communications},
  volume={12},
  number={1},
  pages={7234},
  year={2021},
  publisher={Nature Publishing Group UK London}
}

@article{yang2025bistable,
  title={Bistable soft shells for programmable mechanical logic},
  author={Yang, Nan and Lan, Yuming and Zhao, Miao and Shi, Xiaofei and Huang, Kunpeng and Mao, Zhongfa and Padovani, Damiano},
  journal={Advanced Science},
  volume={12},
  number={5},
  pages={2412372},
  year={2025},
  publisher={Wiley Online Library}
}

@article{yue2024flexibly,
  title={A Flexibly Function-Oriented Assembly Mechanical Metamaterial},
  author={Yue, Chengbin and Zhao, Wei and Li, Fengfeng and Li, Bingxun and Liu, Liwu and Liu, Yanju and Leng, Jinsong},
  journal={Advanced Functional Materials},
  volume={34},
  number={32},
  pages={2316181},
  year={2024},
  publisher={Wiley Online Library}
}

@article{el2021digital,
  title={Digital logic gates in soft, conductive mechanical metamaterials},
  author={El Helou, Charles and Buskohl, Philip R and Tabor, Christopher E and Harne, Ryan L},
  journal={Nature communications},
  volume={12},
  number={1},
  pages={1633},
  year={2021},
  publisher={Nature Publishing Group UK London}
}

@article{nick2020liquid,
  title={Liquid metal microchannels as digital sensors in mechanical metamaterials},
  author={Nick, Zachary H and Tabor, Christopher E and Harne, Ryan L},
  journal={Extreme Mechanics Letters},
  volume={40},
  pages={100871},
  year={2020},
  publisher={Elsevier}
}

@article{hyatt2023programming,
  title={Programming metastable transition sequences in digital mechanical materials},
  author={Hyatt, Lance P and Harne, Ryan L},
  journal={Extreme Mechanics Letters},
  volume={59},
  pages={101975},
  year={2023},
  publisher={Elsevier}
}

@article{zhang2023meta,
  title={Meta-mechanotronics for self-powered computation},
  author={Zhang, Qianyun and Barri, Kaveh and Jiao, Pengcheng and Lu, Wenyun and Luo, Jianzhe and Meng, Wenxuan and Wang, Jiajun and Hong, Luqin and Mueller, Jochen and Wang, Zhong Lin and others},
  journal={Materials Today},
  volume={65},
  pages={78--89},
  year={2023},
  publisher={Elsevier}
}

@article{hauser2011towards,
  title={Towards a theoretical foundation for morphological computation with compliant bodies},
  author={Hauser, Helmut and Ijspeert, Auke J and F{\"u}chslin, Rudolf M and Pfeifer, Rolf and Maass, Wolfgang},
  journal={Biological cybernetics},
  volume={105},
  pages={355--370},
  year={2011},
  publisher={Springer}
}

@article{hauser2012role,
  title={The role of feedback in morphological computation with compliant bodies},
  author={Hauser, Helmut and Ijspeert, Auke J and F{\"u}chslin, Rudolf M and Pfeifer, Rolf and Maass, Wolfgang},
  journal={Biological cybernetics},
  volume={106},
  pages={595--613},
  year={2012},
  publisher={Springer}
}

@article{nakajima2014exploiting,
  title={Exploiting short-term memory in soft body dynamics as a computational resource},
  author={Nakajima, Kohei and Li, Tao and Hauser, Helmut and Pfeifer, Rolf},
  journal={Journal of The Royal Society Interface},
  volume={11},
  number={100},
  pages={20140437},
  year={2014},
  publisher={The Royal Society}
}

@article{caluwaerts2013locomotion,
  title={Locomotion without a brain: physical reservoir computing in tensegrity structures},
  author={Caluwaerts, Ken and D'Haene, Michiel and Verstraeten, David and Schrauwen, Benjamin},
  journal={Artificial life},
  volume={19},
  number={1},
  pages={35--66},
  year={2013},
  publisher={MIT Press One Rogers Street, Cambridge, MA 02142-1209, USA journals-info~…}
}

@article{bhovad2021physical,
  title={Physical reservoir computing with origami and its application to robotic crawling},
  author={Bhovad, Priyanka and Li, Suyi},
  journal={Scientific Reports},
  volume={11},
  number={1},
  pages={13002},
  year={2021},
  publisher={Nature Publishing Group UK London}
}

@article{eder2018morphological,
  title={Morphological computation-based control of a modular, pneumatically driven, soft robotic arm},
  author={Eder, Michael and Hisch, Florian and Hauser, Helmut},
  journal={Advanced Robotics},
  volume={32},
  number={7},
  pages={375--385},
  year={2018},
  publisher={Taylor \& Francis}
}

@article{tayama2025development,
  title={Development of a Soft Robot with Locomotion Mechanism and Physical Reservoir Computing for Mimicking Gastropods},
  author={Tayama, Yoshimune and Furukawa, Hidemitsu and Ogawa, Jun},
  journal={Journal of Robotics and Mechatronics},
  volume={37},
  number={1},
  pages={105--113},
  year={2025},
  publisher={Fuji Technology Press Ltd.}
}

@article{el2024intelligent,
  title={Intelligent electroactive material systems with self-adaptive mechanical memory and sequential logic},
  author={El Helou, Charles and Hyatt, Lance P and Buskohl, Philip R and Harne, Ryan L},
  journal={Proceedings of the National Academy of Sciences},
  volume={121},
  number={14},
  pages={e2317340121},
  year={2024},
  publisher={National Academy of Sciences}
}

@article{byun2024integrated,
  title={Integrated mechanical computing for autonomous soft machines},
  author={Byun, Junghwan and Pal, Aniket and Ko, Jongkuk and Sitti, Metin},
  journal={Nature Communications},
  volume={15},
  number={1},
  pages={2933},
  year={2024},
  publisher={Nature Publishing Group UK London}
}

@article{yoshimura2024research,
  title={Research on tactile sensation by physical reservoir computing with a robot arm and a Ag2S reservoir},
  author={Yoshimura, Kaiki and Hasegawa, Tsuyoshi},
  journal={Japanese journal of applied physics},
  volume={63},
  number={3},
  pages={03SP17},
  year={2024},
  publisher={IOP Publishing}
}

@article{tanaka2021flapping,
  title={Flapping-Wing Dynamics as a Natural Detector of Wind Direction},
  author={Tanaka, Kazutoshi and Yang, Shih-Hsin and Tokudome, Yuji and Minami, Yuna and Lu, Yuyao and Arie, Takayuki and Akita, Seiji and Takei, Kuniharu and Nakajima, Kohei},
  journal={Advanced Intelligent Systems},
  volume={3},
  number={2},
  pages={2000174},
  year={2021},
  publisher={Wiley Online Library}
}

@article{chen2024thermal,
  title={Thermal Computing with Mechanical Transistors},
  author={Chen, Huyue and Song, Chao and Wu, Jiahao and Zou, Bihui and Zhang, Zhihan and Hu, Zhiyuan and Zou, An and Wang, Zhaoguang and Cho, Yuljae and Yang, Zhuoqing and others},
  journal={Advanced Functional Materials},
  volume={34},
  number={25},
  pages={2401244},
  year={2024},
  publisher={Wiley Online Library}
}

@article{chen2025advances,
  title={Advances in metamaterials for mechanical computing},
  author={Chen, Boxin and Nam, Jisoo and Kim, Miso},
  journal={APL Electronic Devices},
  volume={1},
  number={2},
  year={2025},
  publisher={AIP Publishing}
}

@article{pitti2025informational,
  title={Informational Embodiment: Computational role of information structure in codes and robots},
  author={Pitti, Alexandre and Austin, Max and Nakajima, Kohei and Kuniyoshi, Yasuo},
  journal={Physics of Life Reviews},
  year={2025},
  publisher={Elsevier}
}

@article{muller2017morphological,
  title={What is morphological computation? On how the body contributes to cognition and control},
  author={M{\"u}ller, Vincent C and Hoffmann, Matej},
  journal={Artificial life},
  volume={23},
  number={1},
  pages={1--24},
  year={2017},
  publisher={MIT Press One Rogers Street, Cambridge, MA 02142-1209, USA journals-info~…}
}

@article{hauser2021physical,
  title={Physical reservoir computing in robotics},
  author={Hauser, Helmut},
  journal={Reservoir Computing: Theory, Physical Implementations, and Applications},
  pages={169--190},
  year={2021},
  publisher={Springer}
}

@article{wang2025proprioceptive,
  title={Proprioceptive and Exteroceptive Information Perception in a Fabric Soft Robotic Arm via Physical Reservoir Computing with Minimal Training Data},
  author={Wang, Jun and Qiao, Zhi and Zhang, Wenlong and Li, Suyi},
  journal={Advanced Intelligent Systems},
  volume={7},
  number={4},
  pages={2400534},
  year={2025},
  publisher={Wiley Online Library}
}

@inproceedings{horii2021physical,
  title={Physical reservoir computing in a soft swimming robot},
  author={Horii, Yuta and Inoue, Katsuma and Nishikawa, Satoshi and Nakajima, Kohei and Niiyama, Ryuma and Kuniyoshi, Yasuo},
  booktitle={ALIFE 2021: The 2021 Conference on Artificial Life},
  year={2021},
  organization={MIT Press}
}

@article{chen2018harnessing,
  title={Harnessing bistability for directional propulsion of soft, untethered robots},
  author={Chen, Tian and Bilal, Osama R and Shea, Kristina and Daraio, Chiara},
  journal={Proceedings of the National Academy of Sciences},
  volume={115},
  number={22},
  pages={5698--5702},
  year={2018},
  publisher={National Academy of Sciences}
}

@article{raney2016stable,
  title={Stable propagation of mechanical signals in soft media using stored elastic energy},
  author={Raney, Jordan R and Nadkarni, Neel and Daraio, Chiara and Kochmann, Dennis M and Lewis, Jennifer A and Bertoldi, Katia},
  journal={Proceedings of the National Academy of Sciences},
  volume={113},
  number={35},
  pages={9722--9727},
  year={2016},
  publisher={National Academy of Sciences}
}

@article{rothemund2018soft,
  title={A soft, bistable valve for autonomous control of soft actuators},
  author={Rothemund, Philipp and Ainla, Alar and Belding, Lee and Preston, Daniel J and Kurihara, Sarah and Suo, Zhigang and Whitesides, George M},
  journal={Science Robotics},
  volume={3},
  number={16},
  pages={eaar7986},
  year={2018},
  publisher={American Association for the Advancement of Science}
}

@article{patel2023highly,
  title={Highly dynamic bistable soft actuator for reconfigurable multimodal soft robots},
  author={Patel, Dinesh K and Huang, Xiaonan and Luo, Yichi and Mungekar, Mrunmayi and Jawed, M Khalid and Yao, Lining and Majidi, Carmel},
  journal={Advanced Materials Technologies},
  volume={8},
  number={2},
  pages={2201259},
  year={2023},
  publisher={Wiley Online Library}
}

@article{huang2022bistable,
  title={Bistable programmable origami based soft electricity generator with inter-well modulation},
  author={Huang, Cenling and Tan, Ting and Wang, Zhemin and Nie, Xiaochun and Zhang, Shimin and Yang, Fengpeng and Lin, Zhiliang and Wang, Benlong and Yan, Zhimiao},
  journal={Nano Energy},
  volume={103},
  pages={107775},
  year={2022},
  publisher={Elsevier}
}

@article{kaufmann2022harnessing,
  title={Harnessing the multistability of kresling origami for reconfigurable articulation in soft robotic arms},
  author={Kaufmann, Joshua and Bhovad, Priyanka and Li, Suyi},
  journal={Soft Robotics},
  volume={9},
  number={2},
  pages={212--223},
  year={2022},
  publisher={Mary Ann Liebert, Inc., publishers 140 Huguenot Street, 3rd Floor New~…}
}

@article{song2021cmos,
  title={CMOS-inspired complementary fluidic circuits for soft robots},
  author={Song, Sukho and Joshi, Sagar and Paik, Jamie},
  journal={Advanced Science},
  volume={8},
  number={20},
  pages={2100924},
  year={2021},
  publisher={Wiley Online Library}
}

@article{decker2022programmable,
  title={Programmable soft valves for digital and analog control},
  author={Decker, Colter J and Jiang, Haihui Joy and Nemitz, Markus P and Root, Samuel E and Rajappan, Anoop and Alvarez, Jonathan T and Tracz, Jovanna and Wille, Lukas and Preston, Daniel J and Whitesides, George M},
  journal={Proceedings of the National Academy of Sciences},
  volume={119},
  number={40},
  pages={e2205922119},
  year={2022},
  publisher={National Academy of Sciences}
}

@article{tracz2022tube,
  title={Tube-balloon logic for the exploration of fluidic control elements},
  author={Tracz, Jovanna A and Wille, Lukas and Pathiraja, Dylan and Kendre, Savita V and Pfisterer, Ron and Turett, Ethan and Abrahamsson, Christoffer K and Root, Samuel E and Lee, Won-Kyu and Preston, Daniel J and others},
  journal={IEEE Robotics and Automation Letters},
  volume={7},
  number={2},
  pages={5483--5488},
  year={2022},
  publisher={IEEE}
}

@article{conrad20243d,
  title={3D-printed digital pneumatic logic for the control of soft robotic actuators},
  author={Conrad, Stefan and Teichmann, Joscha and Auth, Philipp and Knorr, N and Ulrich, K and Bellin, D and Speck, Thomas and Tauber, Falk J},
  journal={Science robotics},
  volume={9},
  number={86},
  pages={eadh4060},
  year={2024},
  publisher={American Association for the Advancement of Science}
}

@article{jiao2024reprogrammable,
  title={Reprogrammable metamaterial processors for soft machines},
  author={Jiao, Zhongdong and Hu, Zhenhan and Dong, Zeyu and Tang, Wei and Yang, Huayong and Zou, Jun},
  journal={Advanced Science},
  volume={11},
  number={11},
  pages={2305501},
  year={2024},
  publisher={Wiley Online Library}
}

@article{liu2023discriminative,
  title={Discriminative transition sequences of origami metamaterials for mechanologic},
  author={Liu, Zuolin and Fang, Hongbin and Xu, Jian and Wang, Kon-Well},
  journal={Advanced Intelligent Systems},
  volume={5},
  number={1},
  pages={2200146},
  year={2023},
  publisher={Wiley Online Library}
}

@article{bartlett20153d,
  title={A 3D-printed, functionally graded soft robot powered by combustion},
  author={Bartlett, Nicholas W and Tolley, Michael T and Overvelde, Johannes TB and Weaver, James C and Mosadegh, Bobak and Bertoldi, Katia and Whitesides, George M and Wood, Robert J},
  journal={Science},
  volume={349},
  number={6244},
  pages={161--165},
  year={2015},
  publisher={American Association for the Advancement of Science}
}

@article{wehner2016integrated,
  title={An integrated design and fabrication strategy for entirely soft, autonomous robots},
  author={Wehner, Michael and Truby, Ryan L and Fitzgerald, Daniel J and Mosadegh, Bobak and Whitesides, George M and Lewis, Jennifer A and Wood, Robert J},
  journal={nature},
  volume={536},
  number={7617},
  pages={451--455},
  year={2016},
  publisher={Nature Publishing Group UK London}
}

@article{garrad2019soft,
  title={A soft matter computer for soft robots},
  author={Garrad, Martin and Soter, Gabor and Conn, AT and Hauser, Helmut and Rossiter, Jonathan},
  journal={Science Robotics},
  volume={4},
  number={33},
  pages={eaaw6060},
  year={2019},
  publisher={American Association for the Advancement of Science}
}

@article{xu2023soft,
  title={A soft reconfigurable circulator enabled by magnetic liquid metal droplet for multifunctional control of soft robots},
  author={Xu, Yi and Zhu, Jiaqi and Chen, Han and Yong, Haochen and Wu, Zhigang},
  journal={Advanced Science},
  volume={10},
  number={23},
  pages={2300935},
  year={2023},
  publisher={Wiley Online Library}
}

@article{yan2023origami,
  title={Origami-based integration of robots that sense, decide, and respond},
  author={Yan, Wenzhong and Li, Shuguang and Deguchi, Mauricio and Zheng, Zhaoliang and Rus, Daniela and Mehta, Ankur},
  journal={Nature Communications},
  volume={14},
  number={1},
  pages={1553},
  year={2023},
  publisher={Nature Publishing Group UK London}
}

@article{ducarme2025exotic,
  title={Exotic mechanical properties enabled by countersnapping instabilities},
  author={Ducarme, Paul and Weber, Bart and van Hecke, Martin and Overvelde, Johannes TB},
  journal={Proceedings of the National Academy of Sciences},
  volume={122},
  number={16},
  pages={e2423301122},
  year={2025},
  publisher={National Academy of Sciences}
}

@article{matia2023harnessing,
  title={Harnessing nonuniform pressure distributions in soft robotic actuators},
  author={Matia, Yoav and Kaiser, Gregory H and Shepherd, Robert F and Gat, Amir D and Lazarus, Nathan and Petersen, Kirstin H},
  journal={Advanced Intelligent Systems},
  volume={5},
  number={2},
  pages={2200330},
  year={2023},
  publisher={Wiley Online Library}
}

@article{yue2025embodying,
  title={Embodying soft robots with octopus-inspired hierarchical suction intelligence},
  author={Yue, Tianqi and Lu, Chenghua and Tang, Kailuan and Qi, Qiukai and Lu, Zhenyu and Lee, Loong Yi and Bloomfield-Gadȇlha, Hermes and Rossiter, Jonathan},
  journal={Science Robotics},
  volume={10},
  number={102},
  pages={eadr4264},
  year={2025},
  publisher={American Association for the Advancement of Science}
}

@article{li2021light,
  title={Light-powered soft steam engines for self-adaptive oscillation and biomimetic swimming},
  author={Li, Zhiwei and Myung, Nosang Vincent and Yin, Yadong},
  journal={Science Robotics},
  volume={6},
  number={61},
  pages={eabi4523},
  year={2021},
  publisher={American Association for the Advancement of Science}
}

@article{gorissen2020inflatable,
  title={Inflatable soft jumper inspired by shell snapping},
  author={Gorissen, Benjamin and Melancon, David and Vasios, Nikolaos and Torbati, Mehdi and Bertoldi, Katia},
  journal={Science Robotics},
  volume={5},
  number={42},
  pages={eabb1967},
  year={2020},
  publisher={American Association for the Advancement of Science}
}

@article{tang2020leveraging,
  title={Leveraging elastic instabilities for amplified performance: Spine-inspired high-speed and high-force soft robots},
  author={Tang, Yichao and Chi, Yinding and Sun, Jiefeng and Huang, Tzu-Hao and Maghsoudi, Omid H and Spence, Andrew and Zhao, Jianguo and Su, Hao and Yin, Jie},
  journal={Science advances},
  volume={6},
  number={19},
  pages={eaaz6912},
  year={2020},
  publisher={American Association for the Advancement of Science}
}

@article{chi2022snapping,
  title={Snapping for high-speed and high-efficient butterfly stroke--like soft swimmer},
  author={Chi, Yuhan and Hong, Yiyuan and Zhao, Yifan and Li, Yicheng and Yin, Jie},
  journal={Science Advances},
  volume={8},
  number={46},
  pages={eadd3788},
  year={2022},
  publisher={American Association for the Advancement of Science},
  doi={10.1126/sciadv.add3788}
}

@article{faber2020dome,
  title={Dome-Patterned Metamaterial Sheets},
  author={Faber, Jakob A and Udani, Janav P and Riley, Katherine S and Studart, Andr{\'e} R and Arrieta, Andres F},
  journal={Advanced Science},
  volume={7},
  number={22},
  pages={2001955},
  year={2020},
  publisher={Wiley Online Library}
}

@article{le2019filtered,
  title={Filtered mechanosensing using snapping composites with embedded mechano-electrical transduction},
  author={Le Ferrand, Hortense and Studart, Andr{\'e} R and Arrieta, Andres F},
  journal={ACS nano},
  volume={13},
  number={4},
  pages={4752--4760},
  year={2019},
  publisher={ACS Publications}
}

@inproceedings{thuruthel2020bistable,
  title={A bistable soft gripper with mechanically embedded sensing and actuation for fast grasping},
  author={Thuruthel, Thomas George and Abidi, Syed Haider and Cianchetti, Matteo and Laschi, Cecilia and Falotico, Egidio},
  booktitle={2020 29th IEEE International Conference on Robot and Human Interactive Communication (RO-MAN)},
  pages={1049--1054},
  year={2020},
  organization={IEEE}
}

@article{ramachandran2016elastic,
  title={Elastic instabilities of a ferroelastomer beam for soft reconfigurable electronics},
  author={Ramachandran, V. and Bartlett, M. D. and Wissman, J. and Majidi, C.},
  journal={Extreme Mechanics Letters},
  volume={9},
  pages={282--290},
  year={2016},
  publisher={Elsevier},
  doi={10.1016/j.eml.2016.07.004}
}

@article{holmes2013control,
  title={Control and manipulation of microfluidic flow via elastic deformations},
  author={Holmes, Douglas P. and Tavakol, Behrouz and Froehlicher, Guillaume and Stone, Howard A.},
  journal={Soft Matter},
  volume={9},
  number={29},
  pages={7049--7053},
  year={2013},
  publisher={Royal Society of Chemistry},
  doi={10.1039/C3SM51002F}
}

@article{yang2010thermopneumatically,
  title={A thermopneumatically actuated bistable microvalve},
  author={Yang, B. and Wang, B. and Schomburg, W. K.},
  journal={Journal of Micromechanics and Microengineering},
  volume={20},
  number={9},
  pages={095024},
  year={2010},
  publisher={IOP Publishing},
  doi={10.1088/0960-1317/20/9/095024}
}

@inproceedings{maffli2013mm,
  title={Mm-size bistable zipping dielectric elastomer actuators for integrated microfluidics},
  author={Maffli, L. and Rosset, S. and Shea, H. R.},
  booktitle={Electroactive Polymer Actuators and Devices (EAPAD) 2013},
  volume={8687},
  pages={86872M},
  year={2013},
  organization={SPIE},
  doi={10.1117/12.2009367}
}

@article{zhao2025modular,
  title={Modular chiral origami metamaterials},
  author={Zhao, Tuo and Dang, Xiangxin and Manos, Konstantinos and Zang, Shixi and Mandal, Jyotirmoy and Chen, Minjie and Paulino, Glaucio H},
  journal={Nature},
  volume={640},
  number={8060},
  pages={931--940},
  year={2025},
  publisher={Nature Publishing Group UK London}
}

@article{chen2025non,
  title={A non-electrical pneumatic hybrid oscillator for high-frequency multimodal robotic locomotion},
  author={Chen, Genliang and Long, Yongzhou and Yao, Siyue and Tang, Shujie and Luo, Junjie and Wang, Hao and Zhang, Zhuang and Jiang, Hanqing},
  journal={Nature Communications},
  volume={16},
  number={1},
  pages={1449},
  year={2025},
  publisher={Nature Publishing Group UK London}
}

@article{wang2018toward,
  title={Toward perceptive soft robots: Progress and challenges},
  author={Wang, Hongbo and Totaro, Massimo and Beccai, Lucia},
  journal={Advanced science},
  volume={5},
  number={9},
  pages={1800541},
  year={2018},
  publisher={Wiley Online Library}
}

@article{el2020soft,
  title={Soft actuators for soft robotic applications: A review},
  author={El-Atab, Nazek and Mishra, Rishabh B and Al-Modaf, Fhad and Joharji, Lana and Alsharif, Aljohara A and Alamoudi, Haneen and Diaz, Marlon and Qaiser, Nadeem and Hussain, Muhammad Mustafa},
  journal={Advanced Intelligent Systems},
  volume={2},
  number={10},
  pages={2000128},
  year={2020},
  publisher={Wiley Online Library}
}

@article{cianchetti2018biomedical,
  title={Biomedical applications of soft robotics},
  author={Cianchetti, Matteo and Laschi, Cecilia and Menciassi, Arianna and Dario, Paolo},
  journal={Nature Reviews Materials},
  volume={3},
  number={6},
  pages={143--153},
  year={2018},
  publisher={Nature Publishing Group UK London}
}

@article{li2021self,
  title={Self-powered soft robot in the Mariana Trench},
  author={Li, Guorui and Chen, Xiangping and Zhou, Fanghao and Liang, Yiming and Xiao, Youhua and Cao, Xunuo and Zhang, Zhen and Zhang, Mingqi and Wu, Baosheng and Yin, Shunyu and others},
  journal={Nature},
  volume={591},
  number={7848},
  pages={66--71},
  year={2021},
  publisher={Nature Publishing Group UK London}
}

@article{saranli2001rhex,
  title={RHex: A simple and highly mobile hexapod robot},
  author={Saranli, Uluc and Buehler, Martin and Koditschek, Daniel E},
  journal={The International Journal of Robotics Research},
  volume={20},
  number={7},
  pages={616--631},
  year={2001},
  publisher={SAGE Publications}
}

@article{becker2022active,
  title={Active entanglement enables stochastic, topological grasping},
  author={Becker, Kaitlyn and Teeple, Clark and Charles, Nicholas and Jung, Yeonsu and Baum, Daniel and Weaver, James C and Mahadevan, L and Wood, Robert},
  journal={Proceedings of the National Academy of Sciences},
  volume={119},
  number={42},
  pages={e2209819119},
  year={2022},
  publisher={National Academy of Sciences}
}

@article{blickhan2007intelligence,
  title={Intelligence by mechanics},
  author={Blickhan, Reinhard and Seyfarth, Andre and Geyer, Hartmut and Grimmer, Sten and Wagner, Heiko and G{\"u}nther, Michael},
  journal={Philosophical Transactions of the Royal Society A: Mathematical, Physical and Engineering Sciences},
  volume={365},
  number={1850},
  pages={199--220},
  year={2007},
  publisher={The Royal Society London}
}

@article{mousa2024parallel,
  title={Parallel mechanical computing: Metamaterials that can multitask},
  author={Mousa, Mohamed and Nouh, Mostafa},
  journal={Proceedings of the National Academy of Sciences},
  volume={121},
  number={52},
  pages={e2407431121},
  year={2024},
  publisher={National Academy of Sciences}
}

@article{silva2014performing,
  title={Performing mathematical operations with metamaterials},
  author={Silva, Alexandre and Monticone, Francesco and Castaldi, Giuseppe and Galdi, Vincenzo and Al{\`u}, Andrea and Engheta, Nader},
  journal={Science},
  volume={343},
  number={6167},
  pages={160--163},
  year={2014},
  publisher={American Association for the Advancement of Science}
}

@article{zangeneh2018performing,
  title={Performing mathematical operations using high-index acoustic metamaterials},
  author={Zangeneh-Nejad, Farzad and Fleury, Romain},
  journal={New Journal of Physics},
  volume={20},
  number={7},
  pages={073001},
  year={2018},
  publisher={IOP Publishing}
}

@article{dorin2024embodiment,
  title={Embodiment of parallelizable mechanical logic utilizing multimodal higher-order topological states},
  author={Dorin, Patrick and Wang, Kon-Well},
  journal={International Journal of Mechanical Sciences},
  volume={284},
  pages={109697},
  year={2024},
  publisher={Elsevier}
}

@article{alu2025roadmap,
  title={Roadmap on embodying mechano-intelligence and computing in functional materials and structures},
  author={Al{\`u}, Andrea and Arrieta, Andres F and Del Dottore, Emanuela and Dickey, Michael and Ferracin, Samuele and Harne, Ryan and Hauser, Helmut and He, Qiguang and Hopkins, Jonathan B and Hyatt, Lance P and others},
  journal={Smart Materials and Structures},
  volume={34},
  number={6},
  pages={063501},
  year={2025},
  publisher={IOP Publishing}
}

@article{bilal2017bistable,
  title={Bistable metamaterial for switching and cascading elastic vibrations},
  author={Bilal, Osama R and Foehr, Andr{\'e} and Daraio, Chiara},
  journal={Proceedings of the National Academy of Sciences},
  volume={114},
  number={18},
  pages={4603--4606},
  year={2017},
  publisher={National Academy of Sciences}
}

@article{jaeger2023toward,
  title={Toward a formal theory for computing machines made out of whatever physics offers},
  author={Jaeger, Herbert and Noheda, Beatriz and Van Der Wiel, Wilfred G},
  journal={Nature communications},
  volume={14},
  number={1},
  pages={4911},
  year={2023},
  publisher={Nature Publishing Group UK London}
}

@article{jaeger2021towards,
  title={Towards a generalized theory comprising digital, neuromorphic and unconventional computing},
  author={Jaeger, Herbert},
  journal={Neuromorphic Computing and Engineering},
  volume={1},
  number={1},
  pages={012002},
  year={2021},
  publisher={IOP Publishing}
}

@book{swade2001difference,
  title={Difference engine: Charles Babbage and the quest to build the First Computer},
  author={Swade, Doron and Babbage, Charles},
  year={2001},
  publisher={Viking Penguin}
}

@book{hammack2014albert,
  title={Albert Michelson's Harmonic Analyzer: A Visual Tour of a Nineteenth Century Machine that Performs Fourier Analysis},
  author={Hammack, Bill and Kranz, Steve and Carpenter, Bruce},
  year={2014},
  publisher={Articulate Noise Books}
}

@article{yang2025review,
  title={A Review of Chitosan-Based Electrospun Nanofibers for Food Packaging: From Fabrication to Function and Modeling Insights},
  author={Yang, Ji and Wang, Haoyu and Lou, Lihua and Meng, Zhaoxu},
  journal={Nanomaterials},
  volume={15},
  number={16},
  pages={1274},
  year={2025},
  publisher={MDPI}
}

@article{wang2025re,
  title={Re-Purposing a Modular Origami Manipulator Into an Adaptive Physical Computer for Machine Learning and Robotic Perception},
  author={Wang, Jun and Li, Suyi},
  journal={Advanced Science},
  pages={e09389},
  year={2025},
  publisher={Wiley Online Library}
}

@article{sitti2021physical,
  title={Physical intelligence as a new paradigm},
  author={Sitti, Metin},
  journal={Extreme Mechanics Letters},
  volume={46},
  pages={101340},
  year={2021},
  publisher={Elsevier}
}

@article{rodrigue2017overview,
  title={An overview of shape memory alloy-coupled actuators and robots},
  author={Rodrigue, Hugo and Wang, Wei and Han, Min-Woo and Kim, Thomas JY and Ahn, Sung-Hoon},
  journal={Soft robotics},
  volume={4},
  number={1},
  pages={3--15},
  year={2017},
  publisher={Mary Ann Liebert, Inc. 140 Huguenot Street, 3rd Floor New Rochelle, NY 10801 USA}
}

@article{guo2021review,
  title={Review of dielectric elastomer actuators and their applications in soft robots},
  author={Guo, Yaguang and Liu, Liwu and Liu, Yanju and Leng, Jinsong},
  journal={Advanced Intelligent Systems},
  volume={3},
  number={10},
  pages={2000282},
  year={2021},
  publisher={Wiley Online Library}
}

@article{zhao2022stimuli,
  title={Stimuli-responsive polymers for soft robotics},
  author={Zhao, Yusen and Hua, Mutian and Yan, Yichen and Wu, Shuwang and Alsaid, Yousif and He, Ximin},
  journal={Annual Review of Control, Robotics, and Autonomous Systems},
  volume={5},
  number={1},
  pages={515--545},
  year={2022},
  publisher={Annual Reviews}
}

@article{shen2020stimuli,
  title={Stimuli-responsive functional materials for soft robotics},
  author={Shen, Zequn and Chen, Feifei and Zhu, Xiangyang and Yong, Ken-Tye and Gu, Guoying},
  journal={Journal of Materials Chemistry B},
  volume={8},
  number={39},
  pages={8972--8991},
  year={2020},
  publisher={Royal Society of Chemistry}
}

@article{li2023physical,
  title={Physical artificial intelligence (PAI): the next-generation artificial intelligence},
  author={Li, Yingbo and Li, Zhao and Duan, Yucong and Spulber, Anamaria-Beatrice},
  journal={Frontiers of Information Technology \& Electronic Engineering},
  volume={24},
  number={8},
  pages={1231--1238},
  year={2023},
  publisher={Springer}
}

@article{ijspeert2008central,
  title={Central pattern generators for locomotion control in animals and robots: a review},
  author={Ijspeert, Auke Jan},
  journal={Neural networks},
  volume={21},
  number={4},
  pages={642--653},
  year={2008},
  publisher={Elsevier}
}

@article{mackay2002central,
  title={Central pattern generation of locomotion: a review of the evidence},
  author={MacKay-Lyons, Marilyn},
  journal={Physical therapy},
  volume={82},
  number={1},
  pages={69--83},
  year={2002},
  publisher={Oxford University Press}
}

@article{rajappan2022logic,
  title={Logic-enabled textiles},
  author={Rajappan, Anoop and Jumet, Barclay and Shveda, Rachel A and Decker, Colter J and Liu, Zhen and Yap, Te Faye and Sanchez, Vanessa and Preston, Daniel J},
  journal={Proceedings of the National Academy of Sciences},
  volume={119},
  number={35},
  pages={e2202118119},
  year={2022},
  publisher={National Academy of Sciences}
}

@article{aubin2022towards,
  title={Towards enduring autonomous robots via embodied energy},
  author={Aubin, Cameron A and Gorissen, Benjamin and Milana, Edoardo and Buskohl, Philip R and Lazarus, Nathan and Slipher, Geoffrey A and Keplinger, Christoph and Bongard, Josh and Iida, Fumiya and Lewis, Jennifer A and others},
  journal={Nature},
  volume={602},
  number={7897},
  pages={393--402},
  year={2022},
  publisher={Nature Publishing Group UK London}
}

@article{zangeneh2021analogue,
  title={Analogue computing with metamaterials},
  author={Zangeneh-Nejad, Farzad and Sounas, Dimitrios L and Al{\`u}, Andrea and Fleury, Romain},
  journal={Nature Reviews Materials},
  volume={6},
  number={3},
  pages={207--225},
  year={2021},
  publisher={Nature Publishing Group UK London}
}

@article{qian2025guidance,
  title={A guidance to intelligent metamaterials and metamaterials intelligence},
  author={Qian, Chao and Kaminer, Ido and Chen, Hongsheng},
  journal={Nature Communications},
  volume={16},
  number={1},
  pages={1154},
  year={2025},
  publisher={Nature Publishing Group UK London}
}

@article{tzarouchis2025programmable,
  title={Programmable wave-based analog computing machine: a metastructure that designs metastructures},
  author={Tzarouchis, Dimitrios C and Edwards, Brian and Engheta, Nader},
  journal={Nature Communications},
  volume={16},
  number={1},
  pages={908},
  year={2025},
  publisher={Nature Publishing Group UK London}
}

@article{liu2023cellular,
  title={Cellular automata inspired multistable origami metamaterials for mechanical learning},
  author={Liu, Zuolin and Fang, Hongbin and Xu, Jian and Wang, Kon-Well},
  journal={Advanced Science},
  volume={10},
  number={34},
  pages={2305146},
  year={2023},
  publisher={Wiley Online Library}
}

@article{el2020pressure,
  title={Pressure-Driven Two-Input 3D Microfluidic Logic Gates},
  author={El-Atab, Nazek and Canas, Javier Chavarrio and Hussain, Muhammad M},
  journal={Advanced Science},
  volume={7},
  number={2},
  pages={1903027},
  year={2020},
  publisher={Wiley Online Library}
}

@misc{wang2025embodied,
      title={Embodied multi-modal sensing with a soft modular arm powered by physical reservoir computing}, 
      author={Jun Wang and Suyi Li},
      year={2025},
      eprint={2503.06733},
      archivePrefix={arXiv},
      primaryClass={cs.RO},
      url={https://arxiv.org/abs/2503.06733}, 
}

@INPROCEEDINGS{2013SpineReservoir,
  author={Zhao, Qian and Nakajima, Kohei and Sumioka, Hidenobu and Hauser, Helmut and Pfeifer, Rolf},
  booktitle={2013 IEEE/RSJ International Conference on Intelligent Robots and Systems}, 
  title={Spine dynamics as a computational resource in spine-driven quadruped locomotion}, 
  year={2013},
  volume={},
  number={},
  pages={1445-1451},
  doi={10.1109/IROS.2013.6696539}}

@article{shougat2023self,
  title={A self-sensing shape memory alloy actuator physical reservoir computer},
  author={Shougat, Md Raf E Ul and Kennedy, Scott and Perkins, Edmon},
  journal={IEEE Sensors Letters},
  volume={7},
  number={5},
  pages={1--4},
  year={2023},
  publisher={IEEE}
}

@article{xi2025kinematically,
  title={A kinematically Bifurcated Metamaterial for Integrated Logic Operation and Computing},
  author={Xi, Kaili and Wei, Jingsong and Zhang, Xiao and Ma, Jiayao and You, Zhong and Chen, Changqing and Chen, Yan},
  journal={Advanced Science},
  pages={e09829},
  year={2025},
  publisher={Wiley Online Library}
}

@inproceedings{mahon2019soft, 
 title={Soft robots for extreme environments: Removing electronic control}, 
 author={Mahon, Stephen T and Buchoux, Anthony and Sayed, Mohammed E and Teng, Lijun and Stokes, Adam A}, 
 booktitle={2019 2nd IEEE international conference on soft robotics (RoboSoft)}, 
 pages={782--787}, 
 year={2019}, 
 organization={IEEE} 
}

@article{stanley2024high, 
 title={High-speed fluidic processing circuits for dynamic control of haptic and robotic systems}, 
 author={Stanley, Andrew A and Roby, Erik S and Keller, Sean J}, 
 journal={Science Advances}, 
 volume={10}, 
 number={14}, 
 pages={eadl3014}, 
 year={2024}, 
 publisher={American Association for the Advancement of Science} 
}

@article{louvet2025reprogrammable,
  title={Reprogrammable, In-Materia Matrix-Vector Multiplication with Floppy Modes},
  author={Louvet, Theophile and Omidvar, Parisa and Serra-Garcia, Marc},
  journal={Advanced Intelligent Systems},
  pages={2500062},
  year={2025},
  publisher={Wiley Online Library}
}

@article{chen2025physical,
  title={Physical Intelligence in Small-Scale Robots and Machines},
  author={Chen, Huyue and Sitti, Metin},
  journal={Advanced Materials},
  pages={e10332},
  year={2025},
  publisher={Wiley Online Library}
}

@article{chen2025advancing,
  title={Advancing physical intelligence for autonomous soft robots},
  author={Chen, Chi and Shi, Pengju and Liu, Zixiao and Duan, Sidi and Si, Muqing and Zhang, Chuanwei and Du, Yingjie and Yan, Yichen and White, Timothy J and Kramer-Bottiglio, Rebecca and others},
  journal={Science Robotics},
  volume={10},
  number={102},
  pages={eads1292},
  year={2025},
  publisher={American Association for the Advancement of Science}
}

@article{mengaldo2022concise,
  title={A concise guide to modelling the physics of embodied intelligence in soft robotics},
  author={Mengaldo, Gianmarco and Renda, Federico and Brunton, Steven L and B{\"a}cher, Moritz and Calisti, Marcello and Duriez, Christian and Chirikjian, Gregory S and Laschi, Cecilia},
  journal={Nature Reviews Physics},
  volume={4},
  number={9},
  pages={595--610},
  year={2022},
  publisher={Nature Publishing Group UK London}
}

@article{zolfagharinejad2025analogue,
  title={Analogue speech recognition based on physical computing},
  author={Zolfagharinejad, Mohamadreza and B{\"u}chel, Julian and Cassola, Lorenzo and Kinge, Sachin and Syed, Ghazi Sarwat and Sebastian, Abu and van der Wiel, Wilfred G},
  journal={Nature},
  pages={1--7},
  year={2025},
  publisher={Nature Publishing Group UK London}
}

@article{milana2025physical,
  title={Physical control: A new avenue to achieve intelligence in soft robotics},
  author={Milana, Edoardo and Santina, Cosimo Della and Gorissen, Benjamin and Rothemund, Philipp},
  journal={Science Robotics},
  volume={10},
  number={102},
  pages={eadw7660},
  year={2025},
  publisher={American Association for the Advancement of Science}
}

@article{mcgeer1990passive,
  title={Passive dynamic walking},
  author={McGeer, Tad and others},
  journal={Int. J. Robotics Res.},
  volume={9},
  number={2},
  pages={62--82},
  year={1990}
}

@article{brown2010universal,
  title={Universal robotic gripper based on the jamming of granular material},
  author={Brown, Eric and Rodenberg, Nicholas and Amend, John and Mozeika, Annan and Steltz, Erik and Zakin, Mitchell R and Lipson, Hod and Jaeger, Heinrich M},
  journal={Proceedings of the National Academy of Sciences},
  volume={107},
  number={44},
  pages={18809--18814},
  year={2010},
  publisher={National Academy of Sciences}
}

@article{reijniers2010morphology,
  title={Morphology-induced information transfer in bat sonar},
  author={Reijniers, Jonas and Vanderelst, Dieter and Peremans, Herbert},
  journal={Physical review letters},
  volume={105},
  number={14},
  pages={148701},
  year={2010},
  publisher={APS}
}

@article{comoretto2025embodying,
  title={Embodying mechano-fluidic memory in soft machines to program behaviors upon interactions},
  author={Comoretto, Alberto and Koppen, Stijn and Mandke, Tanaya and Overvelde, Johannes TB},
  journal={Device},
  volume={3},
  number={10},
  year={2025},
  publisher={Elsevier}
}

@article{yang2025role,
  title={Role of geometric gradients and size effects in multi-shape memory kirigami metamaterials},
  author={Yang, Hang and Qi, Haobo and Pasini, Damiano},
  journal={Structural and Multidisciplinary Optimization},
  volume={68},
  number={12},
  pages={266},
  year={2025},
  publisher={Springer}
}

@article{mousa2024frequency,
  title={Frequency-Controlled Fluidic Oscillators for Soft Robots},
  author={Mousa, Mostafa and Rezanejad, Ashkan and Gorissen, Benjamin and Forte, Antonio E},
  journal={Advanced Science},
  volume={11},
  number={43},
  pages={2408879},
  year={2024},
  publisher={Wiley Online Library}
}

@article{comoretto2025physical,
  title={Physical synchronization of soft self-oscillating limbs for fast and autonomous locomotion},
  author={Comoretto, Alberto and Schomaker, Harmannus AH and Overvelde, Johannes TB},
  journal={Science},
  volume={388},
  number={6747},
  pages={610--615},
  year={2025},
  publisher={American Association for the Advancement of Science}
}

@article{van2024bio,
  title={Bio-inspired autonomy in soft robots},
  author={van Laake, Lucas Carolus and Overvelde, Johannes Tesse Bastiaan},
  journal={Communications Materials},
  volume={5},
  number={1},
  pages={198},
  year={2024},
  publisher={Nature Publishing Group UK London}
}

\bibliographystyle{Science}



\clearpage


\end{document}